\documentclass[journal]{IEEEtran}

\usepackage{cite}
\usepackage{times}
\usepackage{epsfig}
\usepackage{graphicx}
\usepackage{amsmath}
\usepackage{amssymb}
\usepackage{multirow}
\usepackage{multicol}
\usepackage{bm}
\usepackage{bbm}
\usepackage{subcaption}
\usepackage{adjustbox}
\usepackage{listings}
\usepackage{tabularx,booktabs}
\newcolumntype{Y}{>{\centering\arraybackslash}X}
\newcolumntype{Z}{>{\raggedright\arraybackslash}X}

\newcolumntype{L}[1]{>{\raggedright\let\newline\\\arraybackslash\hspace{0pt}}m{#1}}
\newcolumntype{C}[1]{>{\centering\let\newline\\\arraybackslash\hspace{0pt}}m{#1}}
\newcolumntype{R}[1]{>{\raggedleft\let\newline\\\arraybackslash\hspace{0pt}}m{#1}}

\newcommand{\onedot}{.\null}
\newcommand{\etal}{\emph{et al}\onedot}
\def\eg{\emph{e.g}\onedot}
\def\ie{\emph{i.e}\onedot}

\def\etc{\emph{etc}\onedot}
\DeclareMathOperator*{\argmax}{argmax}
\DeclareMathOperator*{\argmin}{argmin}

\usepackage[dvipsnames]{xcolor}
\newcommand{\revise}[1]{\textcolor{black}{#1}}
\newcommand{\revisetwo}[1]{\textcolor{black}{#1}}

\usepackage{xstring}
\newcommand{\autoref}[1]{\IfBeginWith{#1}{fig:}{Fig.~\ref{#1}}{\IfBeginWith{#1}{tab:}{TABLE~\ref{#1}}{\IfBeginWith{#1}{eq:}{Eq.~\ref{#1}}{\IfBeginWith{#1}{sec:}{Sec.~\ref{#1}}{}}}}}

\begin{document}

\title{Spatio-Temporal Perturbations for Video Attribution}

\author{Zhenqiang Li,
        Weimin Wang,
        Zuoyue Li,
        Yifei Huang,
        Yoichi Sato,~\IEEEmembership{Senior Member,~IEEE}
\thanks{This work is partially supported by JST AIP Accelerated Program Grant Number JPMJCR20U1, a project commissioned by the New Energy and Industrial Technology Development Organization (NEDO) and JSPS KAKENHI Grant Number JP20H04205. \textit{(Corresponding author: Weimin Wang)}}

\thanks{Zhenqiang Li, Yifei Huang and Yoichi Sato are with the Institute of Industrial Science, The University of Tokyo, Tokyo 1538505, Japan (e-mail: \{lzq, hyf,ysato\}@iis.u-tokyo.ac.jp)}
\thanks{Weimin Wang is with the DUT-RU International School of Information Science and Engineering, Dalian University of
Technology, Dalian, China (e-mail: wangweimin@dlut.edu.cn)}
\thanks{Zuoyue Li is with the Department of Computer Science, ETH Zurich, Zurich 8092, Switzerland (e-mail: li.zuoyue@inf.ethz.ch)}
}

\markboth{IEEE TRANSACTIONS ON CIRCUITS and SYSTEMS FOR VIDEO TECHNOLOGY, Early Access. DOI 10.1109/TCSVT.2021.3081761}
{Shell \MakeLowercase{\textit{et al.}}: Bare Demo of IEEEtran.cls for Journals}

\maketitle

\begin{abstract}
The attribution method provides a direction for interpreting opaque neural networks in a visual way by identifying and visualizing the input regions/pixels that dominate the output of a network. Regarding the attribution method for visually explaining video understanding networks, it is challenging because of the unique spatiotemporal dependencies existing in video inputs and the special 3D convolutional or recurrent structures of video understanding networks. However, most existing attribution methods focus on explaining networks taking a single image as input and a few works specifically devised for video attribution come short of dealing with diversified structures of video understanding networks. In this paper, we investigate a generic perturbation-based attribution method that is compatible with diversified video understanding networks. Besides, we propose a novel regularization term to enhance the method by constraining the smoothness of its attribution results in both spatial and temporal dimensions. In order to assess the effectiveness of different video attribution methods without relying on manual judgement, we introduce reliable objective metrics which are checked by a newly proposed reliability measurement. We verified the effectiveness of our method by both subjective and objective evaluation and comparison with multiple significant attribution methods.
\end{abstract}

\IEEEpeerreviewmaketitle

\makeatletter
\def\ps@IEEEtitlepagestyle{
  \def\@oddfoot{\mycopyrightnotice}
  \def\@evenfoot{}
}
\def\mycopyrightnotice{
  {\footnotesize
  \begin{minipage}{\textwidth}
  \centering
  1051-8215 (c) 2021 IEEE. Personal use of this material is permitted. However, permission to use this material for any other purposes must be obtained from the IEEE by sending a request to pubs-permissions@ieee.org.
  \end{minipage}
  }
}

\section{Introduction}
\label{sec:introduction}

Deep neural networks have achieved remarkable performance in various video understanding tasks such as action recognition~\cite{two_stream,i3d,tsn,tcsvt_action_recog1}, video captioning~\cite{dense_videocaption,videocap_transf,tcsvt_video_caption}, video question answering~\cite{vqa,memory_vqa,tcsvt_vqa}, video saliency prediction and detection~\cite{tcsvt_saliency_prediction1, tcsvt_saliency_prediction2,tcsvt_saliency_detection1,tcsvt_saliency_detection2} \etc~However, these networks often perform an opaque nature in their inference process. For example, when classifying two videos of swimming and basketball-playing, it is difficult to identify what elements are relied upon by an action recognition model, the scene information in the background, or the actions of performers. \revisetwo{Explaining and understanding black-box deep networks shows significant potential for analyzing failure cases}, improving the model structure design, and even revealing shortcomings in the training data~\cite{zeiler2014visual}.

Since a neural network can be considered as a mapping from the input space to the output space, the task of explaining and understanding the network can be achieved by answering two main questions~\cite{meaningful_perturb}: (1) which part of an input contributes more to the output of the network~\cite{baehrens2010explain,grad}; (2) how the network achieves this mapping through its internal mechanism~\cite{this_looks_like,interp_cnns,guided_zoom}.
Currently, most works concentrate more on solving the first ``which part'' question via the \textbf{input attribution method}~\cite{integrate_grad_icml}, \ie, attributing the output of a network to specific elements in the input. The method assigns a value to each input element to quantify its contribution to the output and arranges these values in the same shape as the input to form heatmaps (also called attribution maps), which provide a visual way to explain networks.

\begin{figure}[t]
    \centerline{\includegraphics[width=1.0\linewidth]{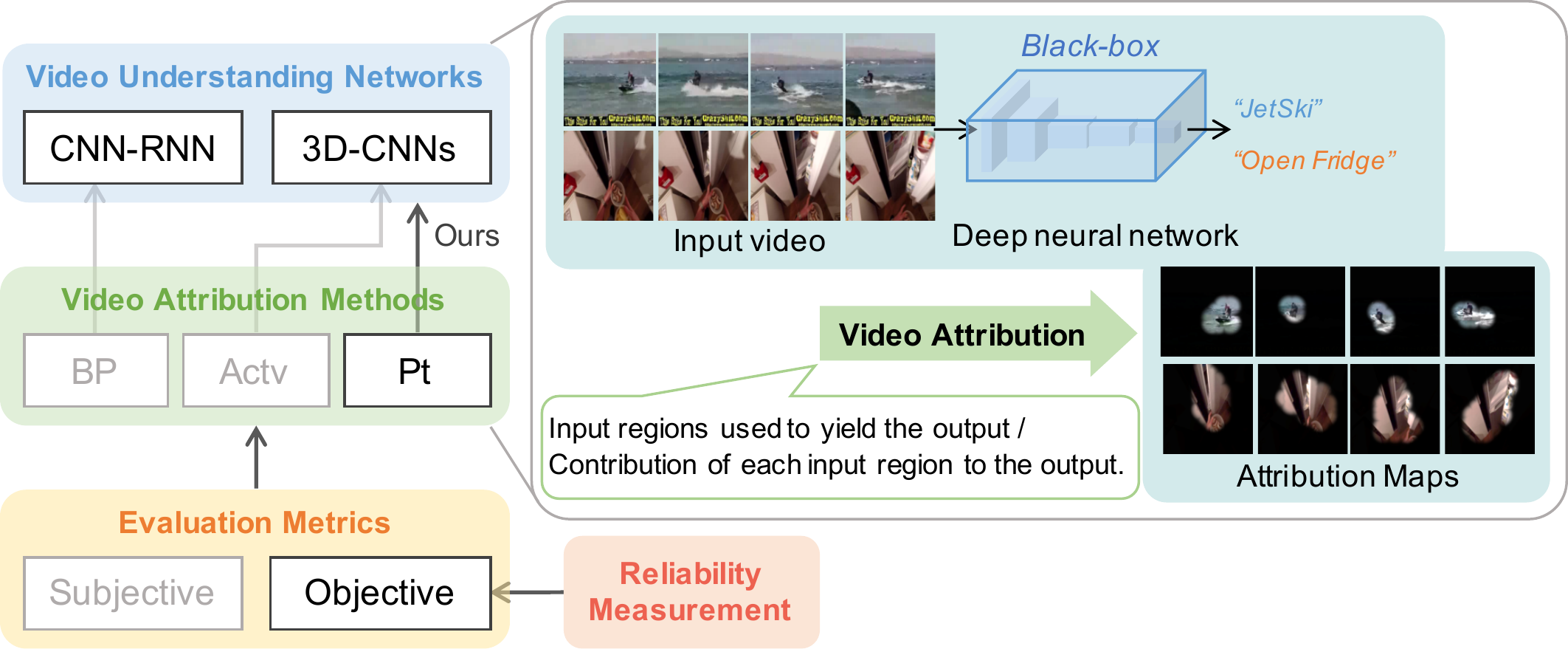}}
    \caption{\textbf{The video attribution task for visually explaining video understanding networks.} We propose a novel Perturbation-based (Pt) method that could be applied on diversified networks such as 3D-CNNs and CNN-RNN because of its model-agnostic characteristic. In contrast, previous video attribution methods based on Backpropagation (BP) or Activation (Actv) may only work well for a specific type of network. Furthermore, for comprehensive evaluations, we incorporate the objective metrics to evaluate video attribution methods in addition to the subjective metrics relying on human judgement. To ensure the utilized objective metrics are reliable, we also devise a measurement for checking the reliability of metrics. As illustrated by the flowchart in the left column, this paper involves the full work stack for the video attribution task.}
    \label{fig:intro}
    \vspace{-2em}
\end{figure}

In this paper, we focus on utilizing the input attribution methods to visually explain video understanding networks. Although input attribution methods have been extensively researched on image recognition networks~\cite{grad,integrate_grad_icml,smoothgrad,lrp,excitbp,grad_cam,meaningful_perturb}, it is nontrivial to investigate attribution methods specifically on video understanding networks because the unique spatiotemporal dependencies in video inputs and the special 3D convolutional or recurrent structures of video understanding networks make it challenging to directly apply existing image attribution methods to the video case. There are a few works~\cite{excit_bp_rnn, saliency_tubes} that focus on the visual explanation for video understanding networks. However, we discover that they still come short of three points: (1) They were designed specifically for only a fixed type of network (\eg 3D-CNNs or CNN-RNN) and cannot be generalized to the diversified networks for video understanding. (2) Their effectiveness was only evaluated by subjective methods, \eg, manual visual inspection or comparison with manual annotations, which deviates from the original intention of the attribute method, \ie, finding the input regions seen important by the network rather than human. (3) They were only compared against a limited number of baseline methods in which some classic and generic attribution methods such as Integrated Gradients~\cite{integrate_grad_icml} and Grad-CAM~\cite{grad_cam} were excluded. An attribution method that is adaptive to the spatiotemporal dependencies in video inputs and generic to diversified video understanding networks is needed but absent. Also, the effectiveness of these methods should be more objectively and comprehensively evaluated.

In response to this demand, we propose a new generic method that is specifically designed for video attribution via leveraging a perturbation-based method and enhancing it by a regularization term for spatiotemporal smoothness constraint. This method inherits the model-agnostic characteristic from the perturbation-based attribution method and is therefore applicable to any video understanding networks without knowing the internal architecture. Furthermore, the new regularization term exploits the spatiotemporal dependencies between frames to generate explanations smoothed in both temporal and spatial dimensions and thus achieves more competitive performance.

In order to assess the effectiveness of different video attribution methods without relying on manual annotation or subjective judgment, we adopt the objective metrics to the video attribution task. Currently, objective evaluation metrics of attribution methods are often established on the perturbation procedure of the input, \ie, sequentially perturbing (inserting/deleting) pixels in the input and quantifying their importance for a network according to the output changes. However, our experiments indicate that different perturbation operations in metrics will yield inconsistent evaluation results. We attribute the reason as the metrics based on the deletion operation are prone to generating adversarial inputs for networks in their calculation process, which results in biased and unreliable evaluation results. Based on this analysis, we propose a new method to measure the reliability of metrics, so that we can select a metric that is able to resist the adversarial effect and produce more reliable evaluation results. Finally, for comprehensive comparison, we bring additional attribution methods as baselines that are competitive in the image attribution task and adaptive for video understanding networks. We compare the effectiveness of our proposed video attribution method with these baseline methods on two typical video understanding backbone networks under the video classification task through both objective and subjective metrics.

This paper extends the preliminary work~\cite{wacv_version} in five major aspects: 
(1) We provide comprehensive evaluations by introducing objective metrics for video attribution methods as supplementary to subjective metrics that rely on human judgment or manual annotations. (2) We devise a new measurement for checking the reliability of different objective evaluation metrics and select metrics that are reliable in the video attribution task based on the assessment. (3) \revise{We introduce more latest baseline methods by adapting multiple typical and generic image attribution methods for the video attribution task. In contrast, in the preliminary work~\cite{wacv_version}, only a limited number of available video attribution methods were compared.} (4) \revisetwo{We add experiments on a new challenging dataset and expand the original two test datasets, which are respectively enlarged 9 and 5 times than what is utilized in the preliminary work~\cite{wacv_version}}. (5) We investigate the influence of the different parameter selections of the newly-proposed regularization term for spatiotemporal perturbations. Our contributions are summarized as follows:
\begin{itemize}
    \item We introduce the perturbation-based method into the video attribution task, which is applicable to diversified and complicated video understanding networks.
    \item We devise a novel regularization term for constraining the spatiotemporal smoothness of the attribution results derived by our perturbation-based method, so as to adapt to the spatiotemporal dependencies in video inputs.
    \item We propose a new method to measure the reliability of objective evaluation metrics for the video attribution methods, which ensures the selected metrics can better resist the adversarial effect and produce more reliable evaluation results.
    \item Both the objective and subjective evaluation results verify that our proposed video attribution method can achieve competitive performances compared with multiple typical and novel attribution methods.
\end{itemize}

\section{Related Work}\label{sec:related_work}
\label{sec:related_work}

In this section, we introduce the literature of input attribution, including existing attribution methods and evaluations.

\begin{table}[t]
    \centering
	\caption[]{\textbf{The summarization and comparison of input attribution methods.} We list typical input attribution methods and compare them from three aspects. The attribution methods that satisfy the three conditions are considered to be applicable as video attribution methods.}
	\label{tab:method_comp}
    \begin{tabularx}{\linewidth}{
                            >{\hsize=0.48\hsize}Z
                            >{\hsize=0.16\hsize}Y
                            >{\hsize=0.2\hsize}Y
                            >{\hsize=0.16\hsize}Y
                        } 
    \toprule
    Attribution Methods & \scriptsize{Unmodified BP Rule} & \scriptsize{All Networks Compatible} & \scriptsize{Unreduced Resolution}\\
    \midrule
    
    \multicolumn{4}{l}{\textbf{Backpropagation-based (BP-based)}} \\
    \cmidrule(lr){1-4}
    Gradient~\cite{grad} & \checkmark & \checkmark & \checkmark \\
    Integrated Gradient~\cite{integrate_grad_icml} & \checkmark & \checkmark & \checkmark \\
    SmoothGrad~\cite{smoothgrad} & \checkmark & \checkmark & \checkmark \\
    Gradient$\times$Input~\cite{linear_appr} & \checkmark & \checkmark & \checkmark \\
    DeConvNets~\cite{zeiler2014visual} & & & \checkmark \\
    Guided BP~\cite{guided_bp_iclr} & & & \checkmark \\
    LRP~\cite{lrp} & & & \checkmark \\
    DeepLift~\cite{deeplift} & & & \checkmark \\
    Excitation BP~\cite{excitbp, excit_bp_rnn} & & & \\
    \revise{SmoothGrad-Squared~\cite{roar}} & \checkmark & \checkmark & \checkmark \\
    \revise{XRAI~\cite{xrai}} & \checkmark & \checkmark & \checkmark \\
    \revise{Blur Integrated Gradient~\cite{blur_ig}} & \checkmark & \checkmark & \checkmark \\
    \midrule

    \multicolumn{4}{l}{\textbf{Activation-based}}   \\
    \cmidrule(lr){1-4}
    CAM~\cite{cam} & \checkmark & & \\
    Grad-CAM~\cite{grad_cam} & \checkmark & & \\
    Grad-CAM++~\cite{grad_cam_plusplus} & \checkmark & & \\
    \revise{Score-CAM~\cite{score_cam}} & \checkmark & & \\
    \midrule

    \multicolumn{4}{l}{\textbf{Perturbation-based}} \\
    \cmidrule(lr){1-4}
    Occlusion~\cite{zeiler2014visual} & \checkmark & \checkmark & \checkmark \\
    LIME~\cite{lime} & \checkmark & \checkmark & \checkmark \\
    RISE~\cite{rise} & \checkmark & \checkmark & \checkmark \\
    Meaningful Perturbation~\cite{meaningful_perturb} & \checkmark & \checkmark & \checkmark \\
    I-GOS~\cite{igos} & & \checkmark & \checkmark \\
    FG-Vis~\cite{fgvis} & & \checkmark & \checkmark \\
    Extremal Perturbation~\cite{extremal_perturb} & \checkmark & \checkmark & \checkmark \\
    \bottomrule
    \end{tabularx}
    \vspace{-1em}
\end{table}

\subsection{Input Attribution Methods}
Given an input and a neural network with fixed parameters, the goal of an input attribution method is to identify the contribution of each element in the input to a specific target output neuron in the network, \eg, the output neuron correlated to the correct class in an image recognition network. The contributions are commonly gathered together to have the same shape as the input and visualized in a form of heatmaps or saliency maps. 
\revisetwo{Similarly, the saliency methods~\cite{tcsvt_saliency_prediction1, tcsvt_saliency_prediction2, tcsvt_saliency_detection1, tcsvt_saliency_detection2} and the networks with attention mechanisms~\cite{tcsvt_action_recognition} can also produce heatmap-like results but the heatmaps have different meanings and goals. The saliency methods aim to localize human-centred salient input regions. The attention mechanism is embedded in a network to enhance its performance by assigning self-learned weights (usually visualized as a heatmap) to different parts of a feature map. In contrast, attribution methods are applied on a pre-trained network with fixed parameters and provide explanations to the network by indicating the contributions of inputs with heatmaps.}

The attribution methods have been extensively researched in previous works and three main types have evolved. Since almost all of them were researched on the image recognition network, we will first summarize these methods according to their types in the case of images by default, and then introduce methods that are especially proposed for videos.

\subsubsection{Backpropagation-based (BP-based) methods}
BP-based attribution approaches are established on a straightforward view that gradients (of the output with respect to the input) could highlight key components in the input since they characterize how much variation would be triggered on the output by a tiny change on the input. Baehrens \etal~\cite{baehrens2010explain} and Simonyan \etal~\cite{grad} have shown the correlation between the pixels' importance and their gradients for a target label. However, the attribution maps generated by raw gradients are typically visually noisy. The ways to overcome this problem could be partitioned into three branches. DeConvNets \cite{zeiler2014visual} and Guided BP \cite{guided_bp_iclr} modify the gradient of the ReLU function by discarding negative values during the back-propagation calculation. Integrated Gradient \cite{integrate_grad_icml} and SmoothGrad \cite{smoothgrad} resist noises by accumulating gradients. LRP \cite{lrp}, DeepLift \cite{deeplift} and Excitation BP \cite{excitbp} employ modified backpropagation rules by leveraging a local approximation or a probabilistic Winner-Take-All process. SmoothGrad-Squared~\cite{roar} achieved improvements of SmoothGrad by adding a square operation. XRAI~\cite{xrai} and Blur Integrated Gradient~\cite{blur_ig} make improvements on Integrated Gradient by incorporating the region-based attribution and blurred input baseline respectively. BP-based methods are often computationally efficient because they need only one forward and backward pass to get attribution maps for the inputs. However, compared with other types of attribution methods, the attribution maps generated by BP-based methods tend to be more noisy and sparse, which makes the contributive regions cannot be evidently highlighted.

\subsubsection{Activation-based methods}
Activation-based attribution approaches generate the explanation by linearly combining the activations of the intermediate layers of a network. Different methods vary in the choice of combining weights. CAM \cite{cam} selects parameters on the fully-connected layer as weights, while Grad-CAM \cite{grad_cam} produces the weight by taking an average of the gradients from the output to the activation. Grad-CAM++ \cite{grad_cam_plusplus} replaces the average pooling layer in Grad-CAM with a weighted pooling operator where the coefficients are calculated by the second derivative. Score-CAM~\cite{score_cam} takes each activation map as a mask for the input and uses the predicted probability of one masked input as the corresponding weight. However, activation-based methods are bound with CNNs and can only generate attribution maps from the intermediate layers. Also, it has been found that the activation-based method tends to produce more meaningful attribution maps at the last convolutional layer~\cite{norm_grad} of CNNs, whose activation is smaller in size than the input. Therefore, the attribution maps generated by activation-based methods are typically lower in resolution than the input.

\subsubsection{Perturbation-based methods}
Perturbation-based attribution methods start from an intuitive assumption that the contributions of certain input elements can be reflected by the changes of the outputs when these elements are removed or preserved only in the input. However, to find the optimal results, theoretically, it is necessary to traverse the elements and their possible combinations in the input and observe their impact on the output. Due to the computational cost of the traversal process, how to obtain an approximated optimal solution faster is the research focus of this problem. Occlusion \cite{zeiler2014visual} and RISE \cite{rise} perturb an image by sliding a grey patch or randomly combining occlusion patches, respectively, and then use changes in the output as weights to sum different patch patterns. LIME \cite{lime} approximates networks into linear models and uses a superpixel based occlusion strategy. Meaningful perturbation \cite{meaningful_perturb} converts the problem to an optimization task of finding a preservation mask that can maximize the output probability under the constraints of preservation ratio and shape smoothness. Real-time saliency \cite{realtime_saliency} learns to predict a perturbation mask with an auxiliary neural network. I-GOS~\cite{igos} introduces integrated gradients instead of the normal gradients to improve the convergence of the optimization process and FG-Vis~\cite{fgvis} incorporates certain restrictions in the optimization process to avoid adversarial results. Extremal Perturbation~\cite{extremal_perturb} factorizes the optimization procedure into two steps to solve the problem of the imbalance between several constraining terms. Most perturbation-based methods characterize model-agnostic since they only access the input and output of a network and require no knowledge or modification of the network's internal structure (except for I-GOS and FG-Vis that need to change the BP rule). However, perturbation-based methods are usually time-consuming because they generate the final results by iteratively adjusting inputs and observing outputs.

\subsubsection{Video Attribution Methods}
Although remarkable progress has been achieved on the image attribution task, it is non-trivial to visually explain the video understanding network by attribution methods. This is mainly because diversified network structures (\eg, 3D-CNNs and CNN-RNN) have been developed on video understanding networks to process the extra temporal dimension in videos. Most previous works only focus on one kind of these structures.  Gan \etal~\cite{devnet} and Anders \etal~\cite{lrp_video} applied the pure gradients and LRP respectively to locate the input regions that are taken important by 3D-CNNs. Grad-CAM \cite{grad_cam} is inherently applicable to 3D-CNNs~\cite{sth_sth}. Stergiou \etal~\cite{saliency_tubes, stergiou2019class} adapted activation-based methods to visualize 3D convolutional networks. For the CNN-RNN structure, EB-R (Excitation BP for RNNs) \cite{excit_bp_rnn} extended the Excitation BP attribution method to adapt to the structure of the RNN. However, to our knowledge, there is still no work comprehensively investigating the performance of input attribution methods on both 3D-CNNs and CNN-RNN by embracing existing attribution methods that are generalizable for the video cases.

In \autoref{tab:method_comp}, we summarize the aforementioned typical input attribution methods and compare them from three aspects: whether they can be utilized without modifying BP rules (Unmodified BP Rule), whether they are compatible with all neural networks instead of some specified structures (All Networks Compatible), and whether they can generate attribution maps with the same resolution of the inputs (Unreduced Resolution). Whether an attribution method satisfies these three conditions determines whether it can be easily transferred to the video attribution task without considering the internal architecture or lowering the spatiotemporal resolution of the attribution result. 

\subsection{Evaluation for Attribution Methods}
Devising evaluation metrics for quantifying the fidelity of an attribution method, \ie, the ability of this attribution method in capturing the true relevant input pixels to the target output, is a vital issue. However, it is challenging since the ground truth map that indicates the true contribution of each input element to the network's target output is hardly obtainable. To overcome the challenge, two kinds of ways have emerged to make the evaluation: subjective and objective ways.

The \textbf{subjective} way relies on human judgment and tends to employ manual visual inspection or bounding boxes that locate the label-related regions. For example, Pointing Game \cite{excitbp} is one of the most commonly used metrics by comparing the attribution maps with manually annotated bounding boxes. However, regions that are contributive for the output of the networks are not necessarily consistent with what is from the human judgment. Thus, this may make subjective evaluations divergent with the aim of fidelity quantification.

As for the \textbf{objective} evaluation, one type of metrics is based upon the input perturbation procedure, in which pixels are inserted or removed in the order decided by the attribution maps. Since they assess the attribution maps by computing the area under the curve (AUC) that plots the change of the target output (\eg, softmax probability), we denote them in short as AUC-based metrics below. Typical AUC-based metrics include Area Over the Perturbation Curve (AOPC)~\cite{aopc} and Causal Metrics (CM)~\cite{rise}. AOPC is measured by removing pixels from the input and has two versions which exploit different removing orders: the Most Relevant First (MoRF) or the Least Relevant First (LeRF). When performing MoRF, pixels assigned with higher attribution values will be removed at first. If the attribution method gives good results, \ie, attribution values are consistent with the real contribution of pixels, the sequential removal of pixels will cause the target output to decrease fast and the final AUC to be small. In contrast, when performing LeRF, a good attribution result will lead to a slow decrease of the target output and thus a large AUC. Different from AOPC, CM only adopts the MoRF procedure but also has two versions according to different perturbation operations to pixels: deletion and insertion. Specifically, the deletion metric is calculated by sequentially removing pixels from the input until the input becomes empty, while the insertion metric performs the opposite procedure. We summarize these metrics in \autoref{tab:metic_comp}. It can be seen that `CM (Deletion)' and `AOPC (MoRF)' are essentially the same metric since they perform the same perturbation operation (deletion) and order (MoRF).

\begin{table}[t]
    \centering
	\caption[]{\textbf{The comparison of four typical objective evaluation metrics.} Differences in the operations and orders taken in the perturbation procedures are considered.}
	\vspace{-0.5em}
	\label{tab:metic_comp}
    \begin{tabularx}{\linewidth}{l *{4}{Y}}
    \toprule
    Evaluation Metrics
    & \multicolumn{2}{c}{Perturbation Operation}  
    & \multicolumn{2}{c}{Perturbation Order}\\
    \cmidrule(lr){2-3} \cmidrule(l){4-5}
    & Insertion & Deletion & MoRF & LeRF \\
    \midrule
    CM (Insertion)~\cite{rise} & \checkmark & & \checkmark & \\ 
    CM (Deletion)~\cite{rise}  & & \checkmark & \checkmark & \\
    AOPC (MoRF)~\cite{aopc}    & & \checkmark & \checkmark & \\ 
    AOPC (LeRF)~\cite{aopc}    & & \checkmark & & \checkmark \\ 
    \bottomrule
    \end{tabularx}
    \vspace{-1em}
\end{table}

Additionally, pixels can be perturbed in different units. AOPC~\cite{aopc} and CM~\cite{rise} delete pixels in the unit of the local neighborhood that is selected as a patch with the shape of 9$\times$9. Instead of a fixed shape, Rieger \etal~\cite{irof} proposed to use superpixels~\cite{slic} as the unit of perturbation.

As a supplementary to the AUC-based metrics, Hooker \etal~\cite{roar} proposed the retrain-based metrics which perturb inputs according to attribution maps by multiple ratios and train the same network from scratch using the perturbed training data. The attribution maps that can obviously reduce/retain the prediction accuracy on the perturbed test dataset are considered to be good. However, retrain-based metrics require extensive computational resources on the retraining. Hence, in this paper, we mainly evaluate video attribution methods objectively by AUC-based metrics and take the retrain-based metric as the complement.



\section{Video Attribution via Perturbations}

\begin{figure*}[!t]
    \centering
    \includegraphics[width=0.7\linewidth]{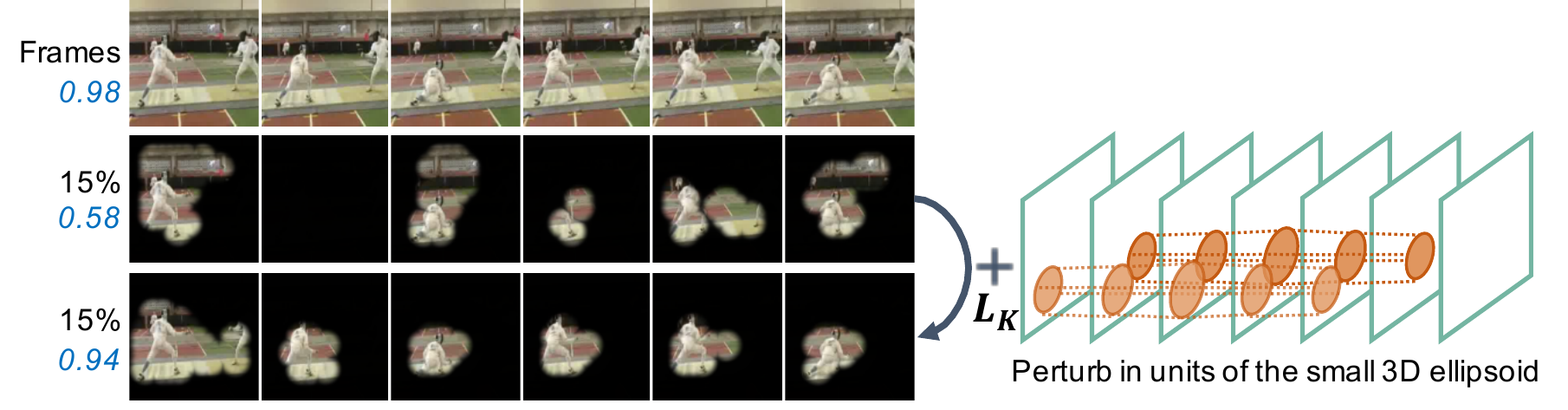}
    \caption[]{\textbf{Illustration of our proposed Spatio-Temporal Extremal Perturbations (STEP) for video attribution.} If we only consider the spatial smoothness in masks, the coherence of the preserved regions in each mask is hard to be ensured and the discriminative regions in the video will also become fragmented (the second row). This makes it difficult to obtain an optimal mask sequence that can retain the output probability (shown as \textcolor{Cerulean}{\textit{0.XX}}). Our method constrains the continuity and shape of the preserved regions in each small spatiotemporal region (illustrated as the orange 3D ellipsoids) through a new regularization term $L_K$, so that the optimized mask sequence can become smoother in both spatial and temporal dimensions and the output probability can be maintained (the third row). X\% denotes the preservation ratio constraint for masks.}
    \label{fig:method_comp}
    \vspace{-1em}
\end{figure*}

\subsection{Perturbation on Videos}

Let $\bm{X}\in\mathbb{R}^{T \times H \times W \times 3}$ represent a video of $T$ frames with width $W$ and height $H$, \revise{RGB $3$ channels,} and $\mathit{\Phi}$ denotes a function that maps the frame sequence to a softmax probability $\mathit{\Phi}_{c}(\bm{X})\in\mathbb{R}$ for a given target class $c$. The goal of video attribution methods is to derive a sequence of attribution maps $\bm{M}\in\left[0,1\right]^{T\times H\times W}$ which assign each pixel $\bm{X}_{t,i,j}$ a value $\bm{M}_{t,i,j}$ that quantifies its contribution to the target output of the function $\mathit{\Phi}_{c}(\bm{X})$. Here $i$ and $j$ refer to the spatial location of each pixel and $t$ refers to the temporal location.

To derive the attribution maps, the key idea of perturbation-based attribution methods is to directly perturb the input to locate the pixels/regions that cause the most significant effects on the output. The \textit{preservation} version of perturbation-based attribution methods~\cite{meaningful_perturb,extremal_perturb} converts this idea to an optimization target that finds a reserving subset of the input which is as small as possible while retaining accuracy on the target output. When applied to video attribution tasks, the optimization problem can be formulated as follows.
\begin{equation}
    \label{eq:ptb_ori_video}
    \bm{M}^*=\argmin_{\bm{M}}\text{ }\{ \lambda\parallel\bm{M}\parallel_{1} - \mathit{\Phi}_c(\bm{M}\otimes{\bm{X}}) \},
\end{equation}
where $\parallel\bm{M}\parallel_{1}:=\sum_{t,i,j}|\bm{M}_{t,i,j}|$ denotes the $L_1$ norm of $\bm{M}$ and $\otimes$ represents the perturbation operation on the input video according to the mask sequence. Specifically, the perturbation is performed independently across each input frame $\bm{X}_t$ and the operation $\otimes$ can be mathematically written as
\begin{equation}
	(\bm{M}\otimes{\bm{X}})_t:=\bm{M}_t\odot\bm{X}_t+(1-\bm{M}_t)\odot(k * \bm{X}_t),
\end{equation}
where $\odot$ denotes the Hadamard multiplication, $*$ represents the 2D convolution, and $k$ is a kernel for Gaussian blur. The first term in \autoref{eq:ptb_ori_video} constrains the preservation ratio on the input video to be small while the second term encourages the model's prediction accuracy to be as high as possible. $\lambda$ controls the balance between the two regularization terms.

However, according to~\cite{extremal_perturb}, due to the difficulty in maintaining the balance between the two constraint targets in \autoref{eq:ptb_ori_video}, it is hard to obtain an optimal solution for this optimization issue. Hence, we adopt the idea of extremal perturbation~\cite{extremal_perturb} and make further adjustments in the case of video inputs. In specific, the two optimization targets in \autoref{eq:ptb_ori_video} is decomposed and arranged to be solved in two steps. The first step finds a binary mask sequence that can maximize the output probability under a constrained preservation ratio $v \in [0,1]$, \ie,
\begin{equation}
    \label{eq:ep_video}
    \bm{M}_v=\argmax_{\bm{M}:\parallel\bm{M}\parallel_1=vTHW} \mathit{\Phi}_c(\bm{M}\otimes\bm{X}),
\end{equation}
while the second step sets the lowest bound $\mathit{\Phi}_{0}$ for the output probability, and searches for the smallest preservation ratio constraint $v^*$ under which the mask sequence can achieve this bound, \ie,
\begin{equation}
    v^*=\min\{v\mid\mathit{\Phi}_c(\bm{M}_v\otimes\bm{X})\geq\mathit{\Phi}_{0}\}.
\end{equation}
Finally, $\bm{M}_{v^*}$ is taken as the final solution.

In order to solve the problem expressed in \autoref{eq:ep_video} with gradient-based optimization approaches, \eg, stochastic gradient descent (SGD), we convert it to a continuous form as below by releasing the binary constraint on the masks,
\begin{equation}
    \label{equ:ep_video_sgd}
    \bm{M}_v=\argmin_{\bm{M}}\{\lambda L_S^v(\bm{M})-\mathit{\Phi}_c(\bm{M}\otimes\bm{X})\},
\end{equation}
\begin{equation}
    \label{eq:ls}
    L_S^v(\bm{M}) = \parallel\text{vecsort}(\bm{M})-\bm{r}_v^{|\bm{M}|}\parallel^2,
\end{equation}
where $\text{vecsort}(\bm{M})\in\mathbb{R}^{THW}$ is a vector that consists of all the elements of $\bm{M}$ sorted in a descending order, $\bm{r}_v^{|\bm{M}|}$ is a template vector which contains $v{|\bm{M}|}$ ones followed by $(1-v){|\bm{M}|}$ zeros, and $|\bm{M}|=THW$. The first regularization term $L_S^v(\bm{M})$ calculates the Euclidean distance between the sorted vector and template vector. It aims to constrain the preservation ratio of $\bm{M}$ to the specified target $v$ through regularizing the values ranked in the top $v{|\bm{M}|}$ to be close to one and the remaining values to approach zero. Then to enforce the preservation ratio on satisfying the constraint as exactly as possible, the weight $\lambda$ is supposed to be large.

\subsection{Spatio-temporal Extremal Perturbations}

It has been discovered that neural networks are vulnerable to adversarial inputs, \eg, images which are unrecognizable to humans might be recognized by networks as arbitrary objects with high confidence~\cite{adversarial_unrecognizable}, or images that are modified in a way imperceptible to humans may mislead networks to have totally wrong predictions~\cite{adversarial_noised}. 
In fact, the optimization target of \autoref{eq:ptb_ori_video} is similar to that for finding adversarial inputs as in \cite{adversarial,adversarial_intriguing,generative_adversarial}. Consequently, 
attribution methods grounded on this optimization target are prone to generating pathological masks, which triggered the adversarial inputs instead of preserving the real contributive input regions~\cite{meaningful_perturb}. 

To avoid these pathological adversarial solutions, Fong \etal~\cite{extremal_perturb} proposed to optimize on smoothed masks that are generated by performing 2D transposed convolution on low-resolution masks with a specially designed 2D smoothing kernel. However, for video attribution methods, the extra temporal dimension in video inputs makes the optimization problem more complex. Although the smoothing technique can improve the optimization result on 2D spatial dimensions, the optimization issue in the temporal dimension remains to be solved for 3D video inputs. The second row of \autoref{fig:method_comp} presents a series of smoothed masks that are generated by solving \autoref{eq:ep_video} ($v=$ 0.15) according to \cite{extremal_perturb}. It can be found that independently smoothing on each 2D mask cannot ensure the temporal coherence of the preserved regions in the mask sequence. Some frames are allocated excessive regions, while others have very few. This uneven and incoherent allocation makes the discriminative spatiotemporal information in a video fragmented, which further makes it difficult to obtain an optimal mask sequence that can retain a high output probability.

Discriminative information in videos commonly continues for a period of time and its corresponding regions will not change sharply in neighbouring frames. Based on this observation, we consider that the preserved regions in a mask sequence should be shaped like tubes that change smoothly in the temporal dimension so as to capture the discriminative regions in a video. However, since the tubes change flexibly according to the information in a video, it is hard to describe their shapes by a fixed mathematical formulation and design a shape regularization term for masks in optimizations. To resolve this challenge, our key insight is that the tubes can be considered to be composed of many small elements with a fixed shape. This means that although we cannot directly regularize the whole shape of the preserved regions, we can constrain the continuity and shape of each preservation in a small spatiotemporal region. Therefore, we design a new regularization term that enforces the high values in masks as concentrated as possible in 3D local neighbourhoods with fixed shapes. Specifically, we implement it by applying 3D convolutions on the attribution maps $\bm{M}$ using a 3D kernel $K$ with a shape of $T_K$, $H_K$, $W_K$ in length, width and height respectively, and then regularize the high values in the convolved masks $\bm{M}* K$ to be close to 1. We mark this new regularization term that constrains the spatiotemporal smoothness of the preserved regions as $L_K$ and mathematically express it as below,
\begin{equation}
    L_K^v(\bm{M}) = L_S^{\beta v}(\bm{M}*K),
\end{equation}
\begin{equation}
    \beta=\frac{{\scriptstyle THW}}{{\scriptstyle(T+T_K-1)(H+H_K-1)(W+W_K-1)}},
\end{equation}
where $*$ denotes 3D convolution with stride and $\beta$ is a scaling factor estimated according to the change of the proportion of 1 in the masks due to the convolution operation.

In our experiment, we set the shape of the small elements that make up the tubes as 3D ellipsoids, considering that they can fit complex tubes more smoothly. Hence, the 3D kernel $K$ is designed to have an ellipsoid shape in which $\forall t\in\{0,...,T_K-1\}, i\in\{0,...,H_K-1\}, j\in\{0,...,W_K-1\}$,
\begin{equation}
    k_{t,i,j}=
    \begin{cases}
      0, &\hspace{-5pt} {\scriptstyle(\frac{2t}{T_K-1}-1)^2+(\frac{2i}{H_K-1}-1)^2+(\frac{2j}{W_K-1}-1)^2 >1}; \\
      1, &\hspace{-5pt} \text{otherwise},
    \end{cases}
\end{equation}
\begin{equation}
    K_{t,i,j}=Z^{-1}k_{t,i,j},
\end{equation}
where $Z=\sum_{t,i,j}k_{t,i,j}$ is the normalization factor. Our experiments indicate that using 3D cylinder can also have a comparable effect. After incorporating the new regularization term $L_K$, \autoref{eq:ep_video} becomes
\begin{equation}
    \label{equ:step_video_sgd}
    \bm{M}_v=\argmin_{\bm{M}}\{
        \lambda_1L_S^v(\bm{M})
        +\lambda_2L_K^v(\bm{M})
        -\mathit{\Phi}_c(\bm{M}\otimes\bm{X})\}.
\end{equation}
Empirically, we set $\lambda_2$ to a value much smaller than $\lambda_1$ in the first few iterations of SGD, and update it to the comparable value as $\lambda_1$ afterward. This can not only fasten the convergence but also ensure the constraint effect of $L_K$ on the shape of the preserved regions in the mask sequence. We call this method for video attribution as \textbf{Spatio-Temporal Extremal Perturbation (STEP)} considering that it can get smoothing extremal perturbation results in both spatial and temporal dimensions. The third row of \autoref{fig:method_comp} shows the masks generated by STEP, in which the preserved regions become smoother in the temporal dimension and the probability is also retained.


\section{Objective Evaluation Metrics}

\subsection{AUC-based Metrics}

The AUC-based evaluation metric is one kind of commonly utilized objective metrics for quantifying the fidelity of one attribution method. Since AUC-based metrics require no retraining of the network, they are suitable for assessing attribution methods on video tasks, whose networks require more extensive computational resources to retrain from scratch.

The calculation of the AUC-based metric is built upon the sequential perturbation procedure of the input. It measures the change of the target output as pixels in the input are sequentially perturbed, and then calculates the area under the curve (AUC) plotting the change. Different AUC-based metrics vary in the perturbation order (Most Relevant First, \textit{abbr}., MoRF or Least Relevant First, \textit{abbr}., LeRF) and perturbation operation (insertion or deletion).

\textbf{MoRF} vs. \textbf{LeRF}: For an input video $\bm{X}$, its pixels' indices set $\bm{I}=\left\{(t,i,j) \mid t=1\dots T,i=1\dots H,j=1\dots W\right\}$ can be divided into $L$ disjoint subsets $\left\{\bm{I}^{(l)}\mid l=1\dots L\right\}$. In the order of MoRF,
pixels with higher attribution values will be perturbed at first, \ie, the aforementioned split should ensure that for each indices pair $\langle(t,i,j), (t^\prime,i^\prime,j^\prime)\rangle$ where $(t,i,j)\in \bm{I}^{(a)}$, $(t^\prime,i^\prime,j^\prime)\in \bm{I}^{(b)}$, and $a<b$, the corresponding values in the attribution maps always satisfy $\bm{M}_{t,i,j}<\bm{M}_{t^\prime,i^\prime,j^\prime}$. On the other hand, if we take LeRF as the perturbation order, pixels with lower attribution values will be perturbed at first and the situation will be reversed.

\textbf{Insertion} vs. \textbf{Deletion}: Pixels in the input are sequentially perturbed in $L$ steps. At the $l^\text{th}$ step, the perturbed input $\tilde{\bm{X}}^{(l)}$ is generated based on a baseline input $\bar{\bm{X}}$ and an incremental perturbation mask sequence $\bm{H}^{(l)}$ as
\begin{equation}
    \tilde{\bm{X}}^{(l)} = \bm{H}^{(l)}\odot\bm{X}+(1-\bm{H}^{(l)})\odot\bar{\bm{X}}.
\end{equation}
We take the baseline input to be the mean of the training data, which satisfies the requirement of $\mathit{\Phi}_{c}(\bar{\bm{X}})\approx0$ for the target class $c$. When performing the insertion operation, the perturbation mask is recursively generated as
\begin{gather}
    \bm{H}^{(0)}={\left\{0\right\}}^{T\times H\times W}, \\
    \bm{H}^{(l)}=\bm{H}^{(l-1)}+\bm{h}^{(l)},
\end{gather}
where $\bm{h}^{(l)}$ satisfies
\begin{equation}
    \bm{h}_{t,i,j}^{(l)}=\mathbbm{1}\left[(t,i,j)\in\bm{I}^{(l)}\right], \forall (t,i,j)\in \bm{I};
\end{equation}
and $\mathbbm{1}\left[\cdot\right]$ denotes the indicator function as 
\begin{equation}
    \mathbbm{1}\left[\textit{cond}\right]=\begin{cases}1 &, \text{if\space} \textit{cond}\text{\space is true}, \\ 0 &, \text{otherwise.}\end{cases}
\end{equation}
For deletion operation, the perturbation mask becomes
\begin{gather}
    \bm{H}^{(0)}={\left\{1\right\}}^{T\times H\times W}, \\
    \bm{H}^{(l)}=\bm{H}^{(l-1)}-\bm{h}^{(l)}.
\end{gather}
Finally the average AUC on a dataset with $N$ input samples is computed as
\begin{equation}
    F=\frac{1}{NL}\sum_{n=1}^{N}\sum_{l=1}^L\mathit{\Phi}_{c}(\tilde{\bm{X}}^{(n)(l)}).
\end{equation}
Here $\tilde{\bm{X}}^{(n)(l)}$ represents the perturbed input of the $n^\text{th}$ sample after the $l^\text{th}$ perturbation step.

\subsection{Reliability Measurement of Metrics}
\label{sec:metric_reliab}

Different versions of the AUC-based metric can be generated by combining different settings such as the perturbation operation and the perturbation order. 
However, in our experiments, we found that the rankings of fidelity evaluations to a group of attribution methods by different versions of the AUC-based metric are inconsistent. Especially, by some metrics, the randomly generated maps tend to have higher fidelity evaluation than what generated by some other attribution methods. This is counterintuitive and implies that the fidelity evaluations obtained by some metrics may be unreliable.

For quantifying the reliability of these metrics, we propose a new measurement. Our measurement mainly focuses on AUC-based metrics, which yield a fidelity evaluation value for the attribute maps on each sample generated by an attribution method. We then designed our reliability measurement based on the following two basic assumptions (axioms):

\textbf{Axiom 1}\label{axiom1}: For an individual input sample, the attribution maps generated according to a reasonable theory have higher fidelity than the maps randomly generated.

\textbf{Axiom 2}\label{axiom2}: For multiple input samples in a dataset, the fidelity rankings for attribution maps generated by a group of attribution methods are consistent across samples.

Assuming there are $M$ attribution methods to be evaluated, and the test dataset contains $N$ samples, then $N\times M$ attribute maps can be obtained. Given one AUC-based metric, we can evaluate the fidelity for all $N\times M$ attribute maps and arrange the fidelity evaluation values into a matrix $\bm{A}\in\mathbb{R}^{N\times M}$, where each row corresponds to values for $M$ attribution maps on one sample and each column contains values for $N$ attribution maps given by one attribution method. Besides, by evaluating the randomly generated results on each sample, a vector $\bm{r}\in\mathbb{R}^N$ containing $N$ values of fidelity can also be obtained. The reliability measurement $\alpha$ of a specific evaluation method can therefore be computed from the result matrix $\bm{A}$ as follows,
\begin{equation}
    \alpha=\frac{\sum_{p=1}^{N-1}\sum_{q=p+1}^{N}w_p w_q\rho(\bm{A}_{p,\cdot},\bm{A}_{q,\cdot})}{\sum_{p=1}^{N-1}\sum_{q=p+1}^{N}w_pw_q},
\end{equation}
where $\rho(\bm{A}_{p,\cdot},\bm{A}_{q,\cdot})$ calculates the Spearman's rank correlation between the fidelity evaluation values for the attribution maps on two different samples. $w_p$ is the ratio of the attribution maps that have better fidelity than the randomly generated maps on the $p^\text{th}$ sample and can be calculated mathematically as below,
\begin{equation}
    w_p=\frac{1}{M}\sum_{m=1}^{M}\mathbbm{1}\left[\bm{A}_{p,m}>r_{p}\right].
\end{equation}
According to Axiom 1, $w_p$ can be used to quantify the reliability score of the evaluation results on one sample, and we use the product of two samples' scores $w_pw_q$ as a weight for the correlation quantification between two samples. The sum of all these products is taken as the normalization factor.

\section{Experiments}\label{sec:experiment}

\subsection{Experiment Setting}
Video classification networks are characterized by complicated and diversified architectures. To compare the results of attribution methods on different video classification networks, we adopt two typical and representative kinds of structures: CNN-RNN and 3D-CNNs. Specifically, we select the ResNet50-LSTM (R50L)~\cite{resnet,lstm} and R(2+1)D~\cite{r2plus1d} models respectively. 
\revise{
We validate attribution methods on three video classification datasets: UCF101-24~\cite{ucf101}, EPIC-Kichens~\cite{epic} and Something-Something-V2 (abbreviated as Sth-Sth-V2)~\cite{sth_sth}. In UCF101-24 and EPIC-Kichens, the manually annotated bounding boxes for the ground-truth labels are available, which are required by the subjective evaluation methods, \eg, pointing game. Sth-Sth-V2 dataset emphasizes the classification of actions from the motion patterns present in human-object interaction instead of the relevant background scenes or static objects. 
}

\subsubsection{Datasets}
\textbf{UCF101-24} is a subset of the UCF101 dataset, containing 3,207 videos of 24 classes that are intensively labeled with spatial bounding box annotations of humans performing actions. In our experiment, we trained models on this dataset using the training split defined in the THUMOS'13 Action Recognition Challenge. When evaluating different attribution methods, we generate attribution maps on the validation set which contains 910 videos.
\textbf{EPIC-Kitchens} is a dataset for egocentric video recognition, where 39,596 video clips segmented from 432 long videos are provided, along with action and object labels. We choose the top 20 object classes with the most number of clips to form the EPIC-Object sub-datasets. 25 clips are randomly selected for each class to generate the validation set and the remaining clips are utilized to train the models. Bounding boxes for the ground-truth objects in EPIC-Objects are provided in 2 fps (one annotation per 30 frames). 
\revise{
\textbf{Sth-Sth-V2} is a video dataset for human-object interaction recognition, which contains 220,847 videos belonging to 174 labels such as `Putting [something] onto [something]'. We construct a sub-dataset by selecting 25 labels with the most number of videos, where 10000 and 1000 videos are picked out for training and validation in our experiments.
}

\subsubsection{Models and training}
We trained an R50L model and an R(2+1)D model on both datasets. In R50L, a deep feature vector for each frame is extracted by ResNet50, which are then temporally accumulated by a one-layer LSTM followed by two fully-connected layers. To alleviate the vanishing of gradients on the beginning input frames, we block the gradient propagation on hidden and cell states of LSTM and take the average of softmax probabilities on all-time steps as the final prediction. Also, we change the activation functions in LSTM from Tanh to ReLU in order to adapt to the requirement of one baseline method~\cite{excit_bp_rnn}. For the R(2+1)D model, we adopt the R(2+1)D-18 structure~\cite{r2plus1d}. In both training and testing phases, we sample 16 frames as the input by splitting one video clip into 16 segments and select one frame from each split. The classification accuracy for each model on every dataset is shown in \autoref{tab:class_acc}. 

\begin{table}[ht]
    \centering
    \caption[]{Top 1 \& 5 classification accuracy of the models for validating video attribution methods. \revisetwo{The asterisk ($*$) indicates the subset of a dataset.} }
	\label{tab:class_acc}
    \begin{tabularx}{\linewidth}{
                            >{\hsize=0.17\hsize}Z |
                            >{\hsize=0.31\hsize}Y
                            >{\hsize=0.31\hsize}Y
                            >{\hsize=0.31\hsize}Y
                        } 
    \toprule
    Accuracy     & UCF101-24 & EPIC-Object$^*$ & Sth-Sth-V2$^*$\\
    \midrule
    R(2+1)D  & 0.97 / 1.00 & 0.71 / 0.94 & 0.66 / 0.90\\
    R50L     & 0.89 / 0.97 & 0.66 / 0.88 & 0.42 / 0.73\\
    \bottomrule
    \end{tabularx}
\end{table}

\subsubsection{Implementation details}
When generating attribution results by our proposed STEP method, all masks are generated based on smaller seed masks $\bar{\bm{M}}\in\mathbb{R}^{\bar{H}\times \bar{W} \times T}$ and in our experiments we set $H=7\bar{H}$ and $W=7\bar{W}$. The seed masks are then up-sampled to $\bm{M}$ by the transposed convolution operation with the 2D smooth max kernel defined in \cite{extremal_perturb}. When performing 3D convolution on the mask sequence for $L_K$, we set the spatial and temporal strides to be 11 and 1 respectively, and padding as 0. We generate a series of masks by choosing 4 area constraints, 0.05, 0.10, 0.15 and 0.20, for both R(2+1)D and R50L. We expect the redundant information can be removed and key regions can be located via small preservation ratios. Empirically, a larger preservation ratio will not arouse a significant increase in the quantitative results. Also, same as \cite{extremal_perturb}, the probability on the ground-truth will saturate after the area constraint exceeding around 20\%. Since masks generated by STEP are nearly binary, to compare with other attribution methods that generate maps with continuous values, we summed these masks and converted the results to heatmaps by applying a Gaussian filter with a standard deviation equals to 10 pixels.

\subsection{Baseline Attribution Methods}
To investigate existing attribution methods and validate the effectiveness of our proposed STEP, we select the following baseline attribution methods and further adapt them for the video inputs.
\begin{itemize}
    \item \textbf{Gradient/Saliency (G)}~\cite{zeiler2014visual} generates the attribution maps based on the gradient of the target output with respect to the input:
    \begin{equation}
        \text{G}(\bm{X})=\frac{\partial\mathit{\Phi}_{c}(\bm{X})}{\partial\bm{X}}.
    \end{equation}
    \item \textbf{Gradient$\times$Input (G*I)}~\cite{linear_appr} extends the gradient method by multiplying the gradients with their corresponding input pixel values:
    \begin{equation}
        \text{G*I}(\bm{X})=\bm{X}\odot\frac{\partial\mathit{\Phi}_{c}(\bm{X})}{\partial\bm{X}}.
    \end{equation}
    \item \textbf{Integrated Gradient (IG)}~\cite{integrate_grad_icml} is defined as the path integral of the gradients along the straight-line path from the baseline $\bar{\bm{X}}$ to the input $\bm{X}$. We compute an approximated version by summing the gradients at a set of $m$ points occurring at small intervals along the straight-line path:
    \begin{equation}
        \text{IG}(\bm{X})=\frac{1}{m}(\bm{X}-\bar{\bm{X}})\odot\sum_{k=1}^{m}\frac{\partial\mathit{\Phi}_{c}(\bar{\bm{X}}+\frac{k}{m}(\bm{X}-\bar{\bm{X}}))}{\partial\bm{X}}.
    \end{equation}
    As suggested by \cite{integrate_grad_icml}, we set $m=$ 50 and use black images as the baseline.
    \item \textbf{SmoothGrad (SG)}~\cite{smoothgrad} calculates attribution maps by averaging the gradients with respect to $n$ noised inputs $\bm{X}^\prime=\bm{X}+\mathcal{N}(0,\sigma^2)$:
    \begin{equation}
        \text{SG}(\bm{X})=\frac{1}{n}\sum_1^n\frac{\partial\mathit{\Phi}_{c}(\bm{X}^{\prime(n)})}{\partial\bm{X}^{\prime(n)}}.
    \end{equation}
    Here $\mathcal{N}(0,\sigma^2)$ represents a 3D tensor having the same shape as $\bm{X}$ with each entry an \textit{i.i.d.} Gaussian noise with 0 mean and standard deviation $\sigma$. We set $n=$ 50 in our experiments.
    \revise{
    \item \textbf{SmoothGrad-Squared (SG2)}~\cite{roar} is a variant of the aforementioned SmoothGrad which squares gradients before averaging them:
    \begin{equation}
        \text{SG2}(\bm{X})=\frac{1}{n}\sum_1^n\Big(\frac{\partial\mathit{\Phi}_{c}(\bm{X}^{\prime(n)})}{\partial\bm{X}^{\prime(n)}}\Big)^2.
    \end{equation}
    Parameter configurations are same as SG.
    \item \textbf{Blur Integrated Gradient (BIG)}~\cite{blur_ig} is a variant of IG which uses the blurred images to replace the black or white images as the baseline. Mathematically, 
    \begin{equation}
        \text{BIG}(\bm{X})=\sum_{k=1}^{m}\frac{\partial\mathit{\Phi}_{c}(g(\bm{X},\sigma_k))}{\partial g(\bm{X},\sigma_k)}\frac{\partial g(\bm{X},\sigma_k)}{\partial \sigma_k}\frac{\sigma_{\text{max}}}{m},
    \end{equation}
    where $g(\bm{X},\sigma_k)$ represents the blurred version of input $\bm{X}$ by a 2D Gaussian kernel with a standard deviation $\sigma_k=(k-1)\sigma_{\text{max}}/m$, $\sigma_{\text{max}}=50$. $m$ is set to $50$ in accordance with IG.
    \item \textbf{XRAI}~\cite{xrai} is a region-based attribution method that builds upon IG. Its key insight is that the region aggregating higher pixel attribution values is more important to the classifier. In our experiments, to adapt to the spatiotemporal continuity of video frames, we employ the SLIC algorithm~\cite{slic} implemented in skimage python package and segment video frames into 3D supervoxels. Same as the original method, we make the segmentation in multiple levels with different parameters (50, 100, 150, 250, 500, 1200) controlling the segment number.
    }
    \item \textbf{Grad-CAM (GC)}~\cite{grad_cam} generates attribution maps from the intermediate layers of a network rather than the input. Specifically, it takes the weighted average of a certain layer's activation $A$ on the different channels as
    \begin{equation}
        \text{GC}(\bm{X})=\text{ReLU}\big(\sum_k\alpha_k A_k\big).
    \end{equation}
    Here $k$ denotes the channel index, and the weight $\alpha_k$ is computed by averaging the gradient with respect to the activation on its spatial dimension.
    \item \textbf{Excitation Backprop (EB)}~\cite{excitbp, excit_bp_rnn} is an attribution method based on a modified back-propagation algorithm that propagates Marginal Winning Probabilities (MWP) in the network. In \cite{excit_bp_rnn}, the back-propagation algorithm is extended to be applicable for RNN. However, since MWP can be only calculated on non-negative neurons, the method can thus only be applied to networks using ReLU as activation functions and generate attribution maps from the intermediate layer.
    
    
\end{itemize}
\vspace{1em}

For GC and EB, we generate attribution results from the last 3D convolutional layer on R(2+1)D and \lstinline{conv4_3} layer on R50L because they have the same spatial size of 7$\times$7. On R(2+1)D, the temporal size of attribution maps generated by GC and EB are eight times smaller than the original frames. For visualization and evaluation, we up-sample their results in both spatial and temporal dimensions. 

\subsection{Metric Reliability Check}

\begin{figure}[b]
    \centering
    \vspace{-2em}
    \includegraphics[width=0.8\linewidth]{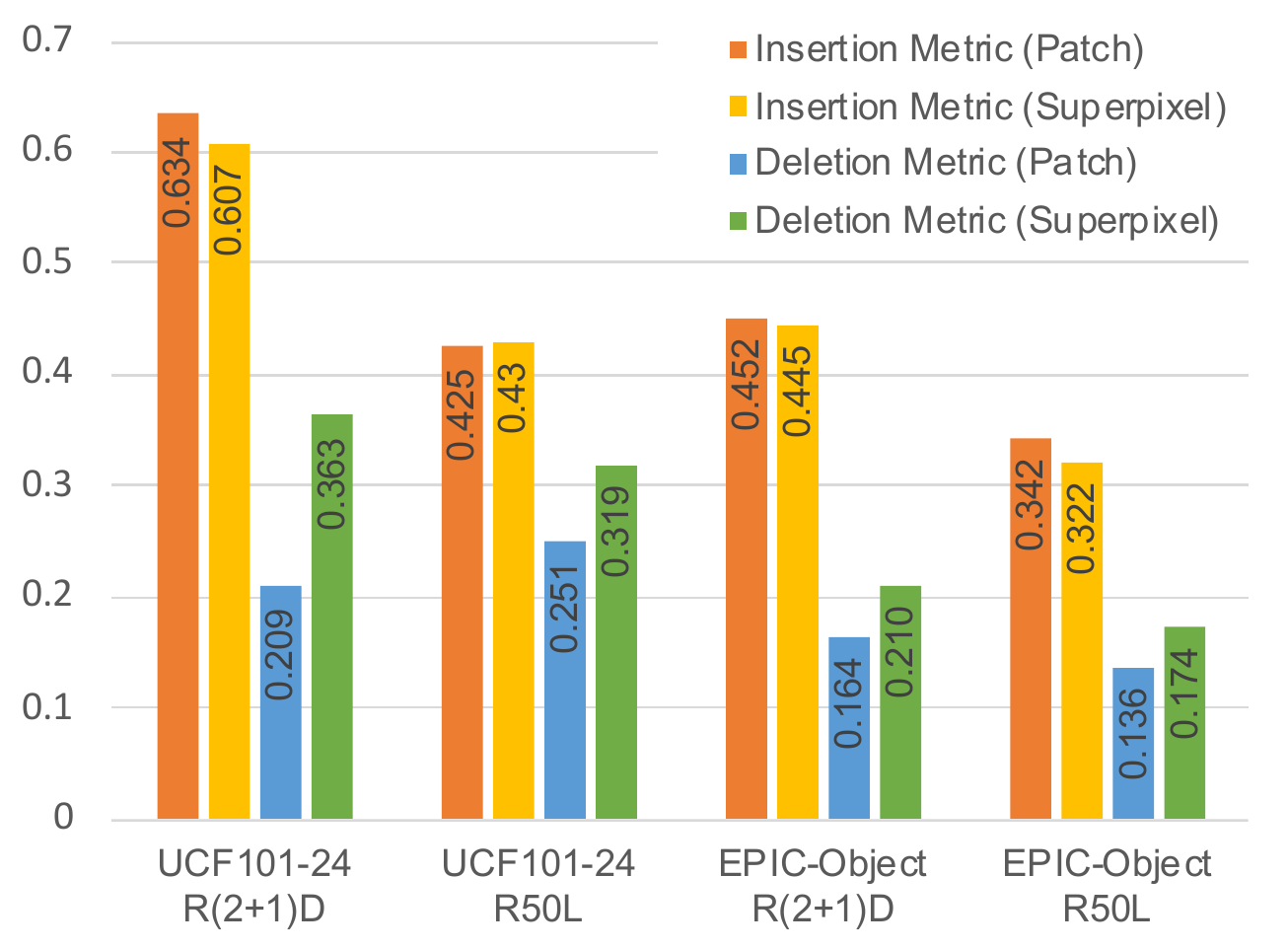} 
    \caption[]{\textbf{Reliability check results of different versions of AUC-based metrics.} The reliability is measured by $\alpha$ as proposed in \autoref{sec:metric_reliab}.}
    \label{fig:metric_check_comp}
\end{figure}

\begin{figure}[th]
    \begin{subfigure}{0.49\linewidth}
        \includegraphics[width=0.95\linewidth]{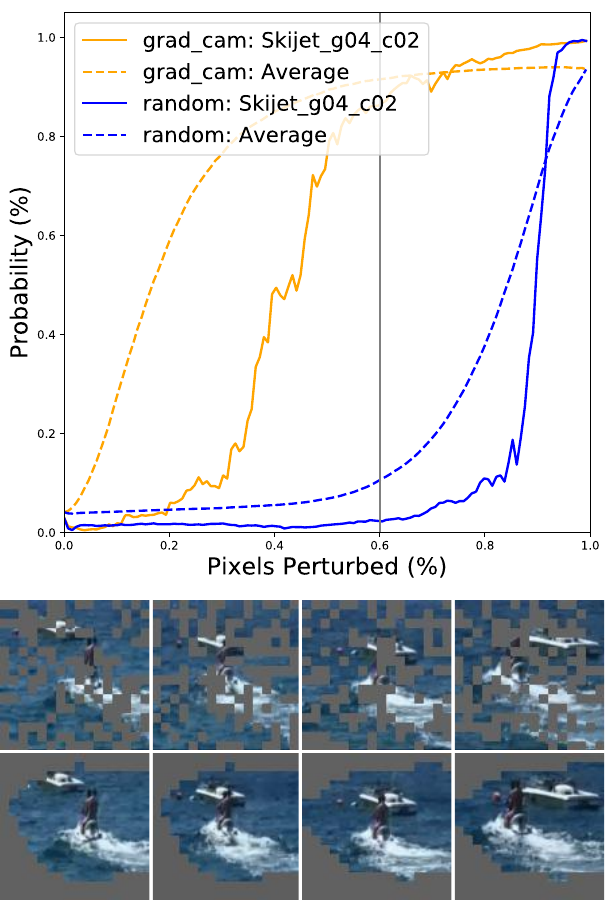} 
        \caption{Insertion Metric}
        \label{fig:metric_check_ins}
    \end{subfigure}
    \begin{subfigure}{0.49\linewidth}
        \includegraphics[width=0.95\linewidth]{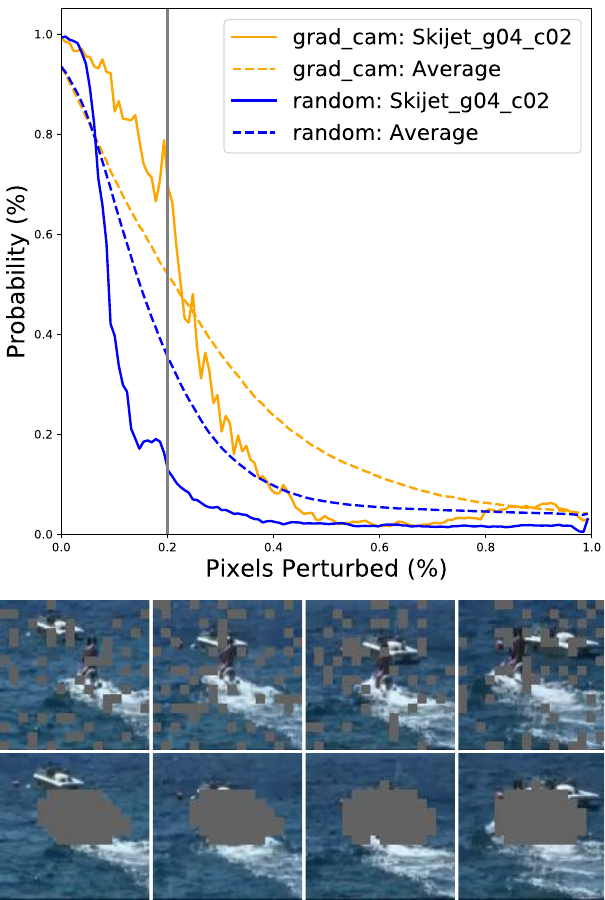} 
        \caption{Deletion Metric}
        \label{fig:metric_check_del}
    \end{subfigure}
    \caption[]{\textbf{Visualization comparison between Insertion and Deletion metrics.} The curves plot the output changes caused by deleting/inserting pixels in the input according to the Grad-CAM attribution maps (\textcolor{Dandelion}{\textbf{---}}) and randomly generated maps (\textcolor{Blue}{\textbf{---}}). Both the average output changes (\textbf{- -}) and a selected sample's output changes (\textbf{---}) are plotted. Also, for each metric, a visualization example for the selected sample is illustrated, in which pixels of a specific ratio (\textcolor{Gray}{$\bm{\mid}$}) are inserted/deleted according to maps.}
    \label{fig:metric_check}
    \vspace{-1em}
\end{figure}

In this paper, we introduce the AUC-based metrics to objectively evaluate the performance of different video attribution methods. AUC-based metrics have evolved out many versions based on the combination of three different variables in the perturbation procedure: operations (insertion/deletion), orders (MoRF/LeRF), and units (Patch/Superpixel). However, based on our observations in experiments, they tend to generate inconsistent evaluations for the same group of attribution methods. Hence, we first check the reliability of different versions based on the measurement we proposed in \autoref{sec:metric_reliab}. For each metric, its reliability measurement is calculated based on the matrix of AUC scores, where each row corresponds to the AUC scores for a set of attribution methods on one input video. In our experiment, we select the random generation and five baseline attribution methods (G, IG, SG, SG2, GC) to form the attribution method set. Because Insertion+MoRF and Deletion+LeRF perform the opposite procedures and produce the equal AUCs (the situation is also the same for Deletion+MoRF and Insertion+LeRF), so we only focus on the two versions under MoRF and denote them as Insertion Metric and Deletion Metric. We set the patch size to be 7$\times$7 when using patch-wise perturbations and split each frame into around 256 segments when perturbing in superpixel. Since the size of input frame is 112$\times$112, this can ensure that the two unit versions are perturbed with approximately the same step size. The reliability check results are shown in \autoref{fig:metric_check_comp}. 

Comparing different perturbation operations, it is obvious that Deletion Metrics show much lower reliability than Insertion Metrics. We consider the reason as the deletion operation is easy to generate inputs that are adversarial for networks, that is, the network output tends to drop dramatically if noncontinuous and scattered regions are removed from an input, although they are small and locate in the unimportant parts. This can be proved by the example illustrated in \autoref{fig:metric_check}, in which the Deletion Metric evaluates the random generation (denoted as Random) to be better than Grad-CAM (shown as the smaller AUC of the averaged output probability). However, when comparing the two visualizations in which the top 20\% pixels are deleted from the input sample according to maps given by the two methods (shown in right-bottom of \autoref{fig:metric_check}), we can find that the input perturbed by Random's maps is more recognizable than that perturbed by Grad-CAM's maps, but their output probabilities show an opposite result. We consider this as an evidence that randomly deleting a small proportion of pixels from the input will cause adversarial inputs.

For different perturbation units, under the insertion operation, the reliability of using Patch and Superpixel are almost the same, while under the deletion operation, Superpixel performs slightly more reliable than Patch. This may because the semantic information in superpixels is more continuous and complete than that in the patchs. Hence, when the discriminative regions in the input are deleted in a form of superpixels, the decline in the output tends to be sharper than that caused by patch-wise deletion and random removal.
Based on the aforementioned analysis, in the following part, we will ignore the Deletion Metric and report the evaluation results for attribute methods under the two versions of the Insertion Metric taking patch and superpixel as the perturbation units.


\subsection{Effect of the New Regularization Term}

To investigate the effectiveness of our proposed regularization term $L_K$ and its influence on the final attribution result of STEP, we first set a group of STEP methods with different kernel lengths $T_K\in\{0,5,8,11,14\}$, and then adopt the Insertion Metric to evaluate the performance of these STEP methods. The evaluation results are shown in \autoref{tab:loss_length}. Comparing the performance of the STEP methods with ($T_K>0$) and without ($T_K=0$) employing the regularization term, we can find that the regularization term can effectively enhance the performance of STEP in most cases.
Especially, the regularization term has the best effect when the kernel length on the temporal dimension is 8. Comparing the results between the two models, it can be found that the enhancement effect of $L_K$ on the R50L model is not significant as that on the R(2+1)D model. We consider that the reason is consistent with our previous analysis, that is, CNN-RNN does not show the same sensitivity as 3D-CNNs to the changes in the input. In the following experiments, we will use the STEP method with $T_K=8$ by default.

\begin{table}[h]
    \centering
    \footnotesize
	\caption[]{\textbf{The influence of $T_K$  in the regularization term $L_K$.} We evaluate STEP methods with different kernel lengths $T_K$ in the regularization term $L_K$ by two versions of Insertion Metric with different perturbation units (Patch/Superpixel). $T_K=0$ means the STEP method without employing $L_K$.}
	
	\label{tab:loss_length}
	\begin{adjustbox}{width=1\linewidth}
    \begin{tabular}{@{}ccccccc@{}}
    \toprule
    \multirow{2}{*}{$T_K$}
    & \multicolumn{2}{c}{UCF101-24}  
    & \multicolumn{2}{c}{EPIC-Object}
    & \multicolumn{2}{c}{Sth-Sth-V2}  \\
    \cmidrule(lr){2-3} \cmidrule(lr){4-5} \cmidrule(l){6-7}
    & R(2+1)D & R50L & R(2+1)D & R50L & R(2+1)D & R50L\\
    \midrule
    0 &  0.761/0.680 & 0.566/0.524 & 0.448/0.409 & \textbf{0.456}/0.467 & 0.355/0.266 & 0.237/0.215 \\
    5 &  0.802/0.735 & 0.575/0.540 & 0.505/0.458 & 0.453/0.473 & 0.376/0.286 & 0.241/0.225  \\
    8 &  \textbf{0.805}/\textbf{0.744} & \textbf{0.575}/\textbf{0.553} & \textbf{0.520}/\textbf{0.484} & 0.455/\textbf{0.487} & 0.389/0.311 & \textbf{0.245}/0.232 \\
    11 & 0.800/0.736 & 0.573/0.551 & 0.509/0.474 & 0.446/0.481 & 0.384/0.307 & 0.242/0.232  \\
    14 & 0.784/0.706 & 0.570/0.537 & 0.495/0.448 & 0.426/0.452 & \textbf{0.393}/0.290 & 0.232/0.216  \\
    \bottomrule
    \end{tabular}
    \end{adjustbox}
    \vspace{-1em}
\end{table}

\subsection{Influence of the Preservation Ratio Constraint}

\begin{figure}[t]
    \centering
    \includegraphics[width=0.9\linewidth]{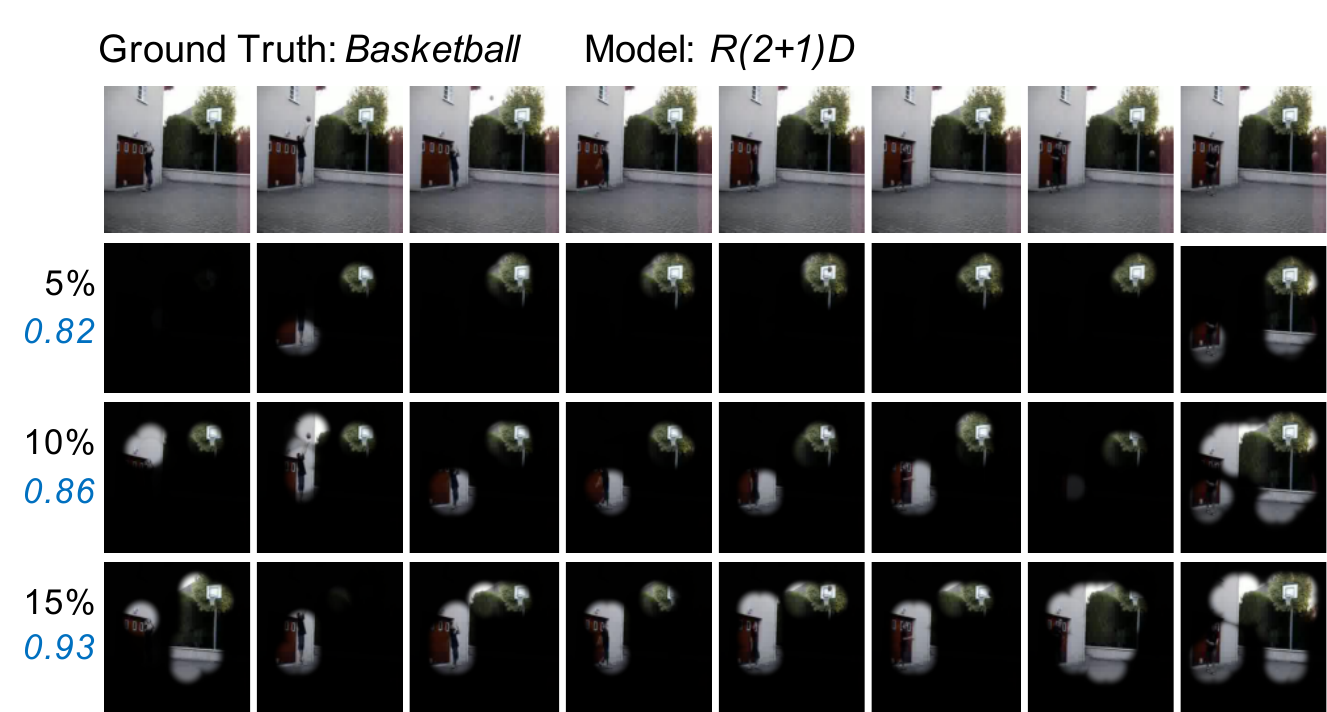}\\
    \includegraphics[width=0.9\linewidth]{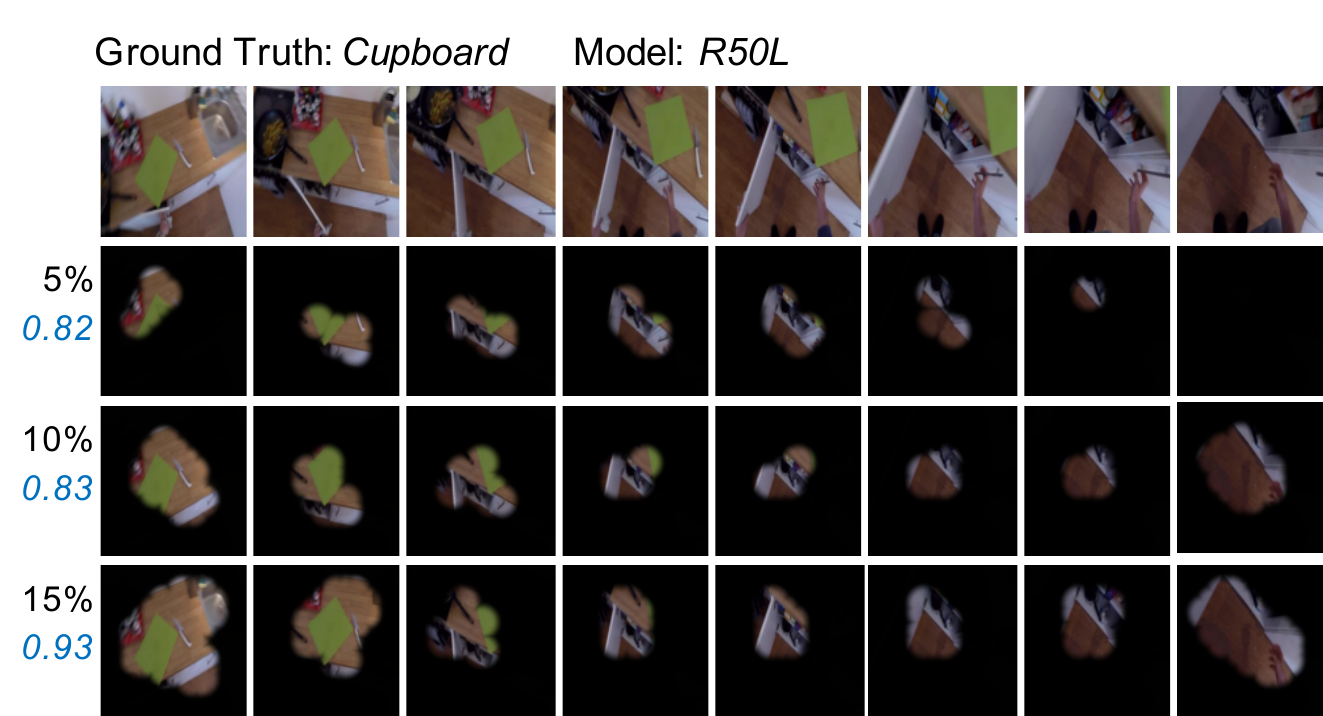}
    \caption[]{\textbf{The attribution masks generated by STEP with different preservation ratio constraints.} They highlight the most discriminative regions for the networks and the predicted probabilities (shown as \textcolor{Cerulean}{\textit{0.XX}}) on the ground-truth labels grow as the preservation ratio increases (shown as X\%).}
    \label{fig:vis_comp_area}
    \vspace{-1em}
\end{figure}

\autoref{fig:vis_comp_area} visualizes the attribution results of STEP with different preservation ratio constraints $v$. The probabilities of predicting the ground-truth label can be high even though only 5\% regions are preserved, which can be considered as the most discriminative regions for the networks. As the preservation ratio constraint increases, more supplementary regions are excavated and the probabilities are promoted further. It is worth noting that on the UCF101-24 dataset, the most discriminative regions may not be considered as the action performer but representative objects in the background (\eg, the backboard for the basketball action). This is reasonable and consistent with some previous observations that networks may make correct predictions by leveraging the bias in a dataset (\eg, scene context or object information) instead of focusing on actual human actions in the videos~\cite{dance_in_mall,diving48}.

\subsection{Qualitative Comparison against Baselines}

\begin{figure*}[th]
    \centering
    \includegraphics[width=0.49\textwidth]{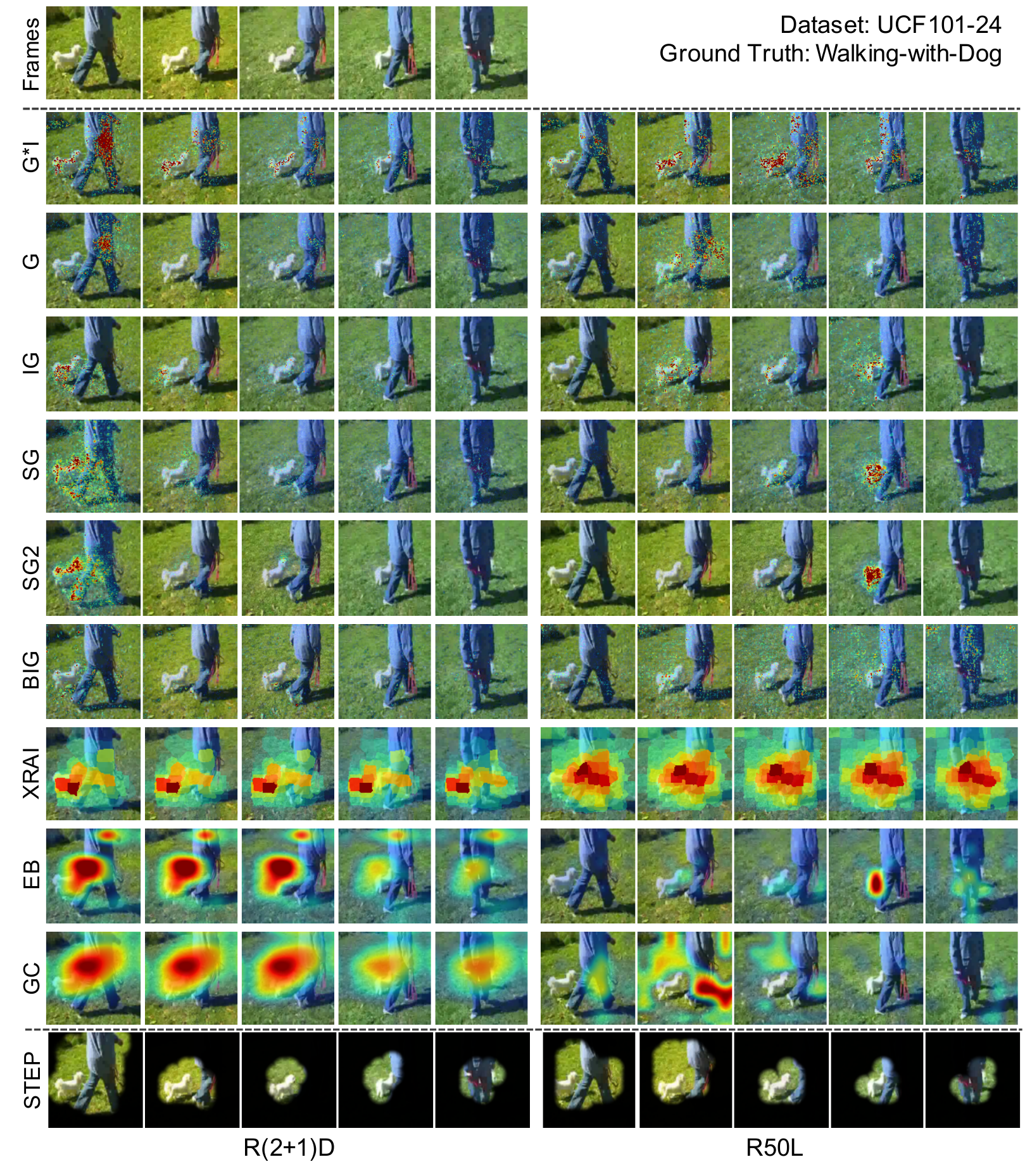}
    \includegraphics[width=0.49\textwidth]{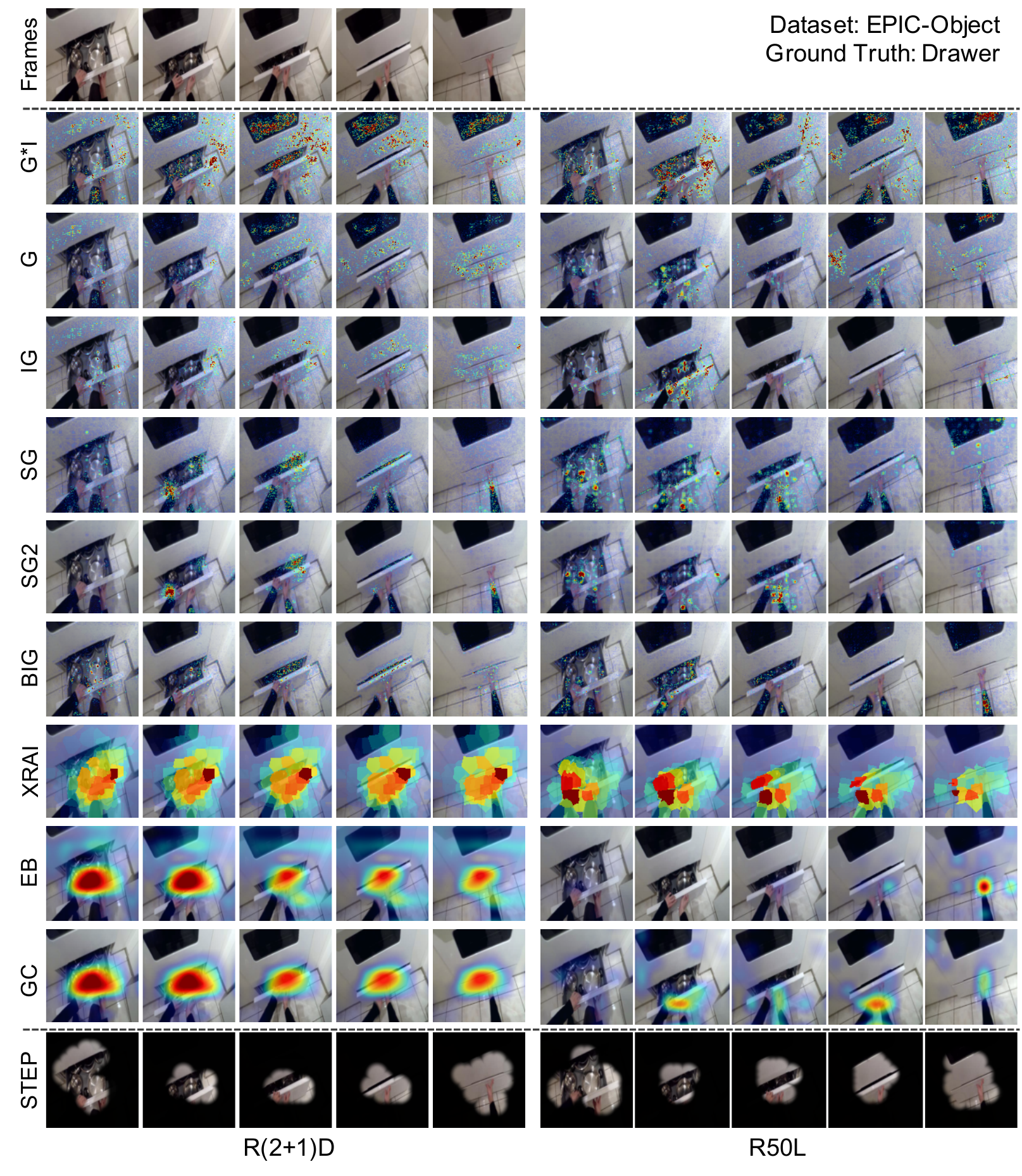}
    \caption[]{\textbf{The visualization of video attribution results and comparison with baseline methods.} We qualitatively compare our proposed STEP methods to baseline video attribution methods: Gradient$\times$Input (G*I)~\cite{linear_appr}, Gradient (G)~\cite{zeiler2014visual}, Integrated Gradient (IG)~\cite{integrate_grad_icml}, SmoothGrad (SG)~\cite{smoothgrad}, SmoothGrad-Squared (SG2)~\cite{roar}, Blur Integrated Gradient (BIG)~\cite{blur_ig}, XRAI~\cite{xrai}, Grad-CAM (GC)~\cite{grad_cam}, Excitation Backprop (EB)~\cite{excitbp,excit_bp_rnn}.}
    \label{fig:vis_comp_baseline}
    \vspace{-1em}
\end{figure*}

\autoref{fig:vis_comp_baseline} illustrates two groups of visualization results including the original frames and attribution maps generated by different video attribution methods. The left group corresponds to a UCF101-24 video with the label of Walking-with-Dog and the right group presents results of a video with the object label of Drawer from the EPIC-Object sub-dataset. 5 frames are sampled out of 16 input frames for visualization. We show the results for the same frame on both R(2+1)D and R50L models. For our STEP methods, we visualize the results generated under the preservation ratio constraint of 0.15. 

It can be seen that our proposed STEP methods can generate maps that are smooth in spatial and sensitive to changes in different video frames. 
For the baseline methods in which gradients are involved, \ie, Gradient$\times$Input, Gradient, Integrated Gradient, SmoothGrad, SmoothGrad-Squared and Blur Integrated Gradient, it is obvious that their attribution results are sparse and noisy in whole, but Integrated Gradient, SmoothGrad, SmoothGrad-Squared can produce relatively concentrated results by exploiting special ways to remove noises. 
In contrast, the region-based attribution method XRAI produces more continuous and smoother results, although the temporal sensitivity is restricted by the supervoxels.
For Grad-CAM and Excitation Backprop that take results from the intermediate layer of networks, the lower original resolution and up-sampling operation make the importance maps looks concentrated and smooth. However, on the R(2+1)D model, the lower temporal resolution would make it inconvenient if we want to have attribution results with high temporal sensitivity. In contrast, our proposed STEP can generate smooth and concentrated attribution results directly from the input, which not only enables the utilization even without the internal structural knowledge of the model but also ensures the sensitivity to changes in different frames.

\subsection{Quantitative Comparison using AUC-based Metrics}

\begin{table}[t]
    \centering
	\caption[]{\textbf{Objective evaluation results for different video attribution methods.} Two versions of the Insertion Metric with different perturbation units (patch/superpixel) are selected for thorough comparisons. The higher value means better performance in attribution.}
	\label{tab:quan}
	
	\begin{adjustbox}{width=1\linewidth}
    \begin{tabular}{@{}lcccccc@{}}
    \toprule
    \multirow{2}{*}{Methods}
    & \multicolumn{2}{c}{UCF101-24}  
    & \multicolumn{2}{c}{EPIC-Object}
    & \multicolumn{2}{c}{\revise{Sth-Sth-V2}}\\
    \cmidrule(lr){2-3} \cmidrule(lr){4-5} \cmidrule(l){6-7}
    & R(2+1)D & R50L & R(2+1)D & R50L & R(2+1)D & R50L \\
    \midrule
    Random & 0.21/0.20 & 0.28/0.27 & 0.17/0.16 & 0.20/0.24 & 0.11/0.10 & 0.12/0.12 \\
    G*I~\cite{linear_appr} & 0.53/0.54 & 0.50/0.49 & 0.36/0.38 & 0.41/0.43 & 0.24/0.24 & 0.20/0.21 \\
    G~\cite{zeiler2014visual} & 0.61/0.60 & 0.51/0.50 & 0.40/0.41 & 0.42/0.45 & 0.25/0.24 & 0.22/0.22 \\
    IG~\cite{integrate_grad_icml} & 0.70/0.68 & 0.54/0.54 & 0.41/0.42 & 0.42/0.46 & 0.31/0.27 & 0.24/0.24 \\
    SG~\cite{smoothgrad} & 0.73/0.70 & 0.55/0.54 & 0.44/0.43 & 0.43/0.46 & 0.29/0.25 & 0.24/0.24 \\
    SG2~\cite{roar} & 0.75/0.73 &  \textbf{0.58}/0.55 & 0.45/0.44 & 0.45/0.47 & 0.30/0.25 & 0.25/\textbf{0.25}\\
    BIG~\cite{blur_ig} & 0.62/0.60 & 0.47/0.49 & 0.39/0.40 & 0.40/0.44 & 0.28/0.24 & 0.21/0.22 \\
    XRAI~\cite{xrai} & 0.74/0.68 & 0.57/\textbf{0.57} & 0.50/0.46 & 0.44/0.45 & 0.50/0.35 & 0.23/0.23 \\
    \midrule
    EB~\cite{excit_bp_rnn} & 0.66/0.59 & 0.57/0.54  & 0.45/0.42 & \textbf{0.46}/0.44 & 0.45/0.31 & 0.24/0.24 \\
    GC~\cite{grad_cam} & 0.74/0.68 & 0.52/0.49 & \textbf{0.56}/\textbf{0.51} & 0.43/0.45 & \textbf{0.54}/\textbf{0.40} & 0.24/0.24 \\
    \midrule
    STEP & \textbf{0.80}/\textbf{0.74} & 0.57/0.55 & 0.52/0.48 & 0.45/\textbf{0.49} & 0.39/0.31 & \textbf{0.25}/0.23 \\
    \bottomrule
    \end{tabular}
    \end{adjustbox}
    \vspace{-1em}
\end{table}

To objectively compare different video attribution methods without relying on manual annotations or biasing to human judgment, we first adopted the AUC-based metric. According to the result of the reliability check \autoref{sec:metric_reliab}, we select the Insertion Metric and evaluate different attribution methods by its two versions which adopt the perturbation unit of patch and superpixel respectively. \autoref{tab:quan} shows the evaluation results and compares our proposed STEP against baseline methods. 

\revise{
We see that our proposed STEP achieves the best evaluation results on multiple columns and competitive performance on the remaining. Our method also maintains good performance on both kinds of networks. Among baseline methods, SG2, XRAI, EB and GC present noticeable performance. It can be found that the attribution methods producing continuous and smooth results can achieve higher scores, especially on R(2+1)D. We attribution this to that these results can generate more continuous inserted regions during the computation procedure of insertion metric, so as to stimulate the network's positive response faster. However, comparing with some methods achieving this continuity and smoothness at the expense of low resolutions (GC and EB) or temporal sensitivity (XRAI), our method keeps a good balance between the two aspects. 
}

\begin{figure}[h]
    \centering
    \includegraphics[width=0.9\linewidth]{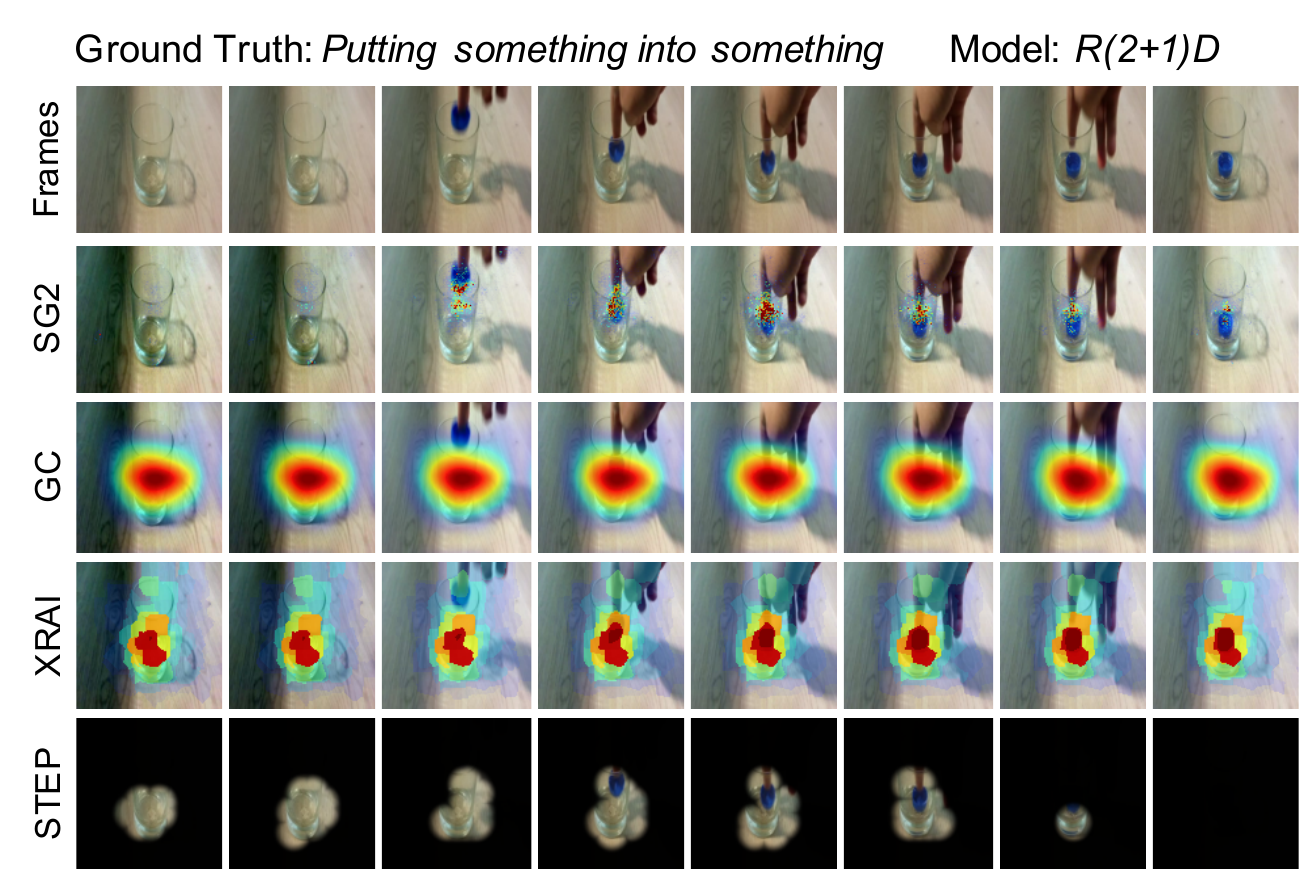}
    \caption[]{\textbf{The comparison of visualization results on R(2+1)D and Something-Something-V2 dataset}. Methods like GC and XRAI achieve high quantitative evaluation results on this setting but show very low temporal sensitivity.}
    \label{fig:vis_sthsthv2_comp}
    \vspace{-1em}
\end{figure}

\revise{
Notably, on the challenging Sth-Sth-V2 dataset that emphasizes the identification of the dynamic motion patterns, GC and XRAI obtain obviously higher evaluation values than other methods although they show low temporal resolution or sensitivity (as shown in \autoref{fig:vis_sthsthv2_comp}). This indicates that on this dataset, differing from our intuition, 3D-CNNs may highly depend on the spatiotemporally continuous regions instead of discrete regions in certain frames to capture the discriminative dynamic information.
}

\subsection{Quantitative Comparison using Pointing Games}

\begin{table}[h]
    \centering
    \footnotesize
	\caption[]{\textbf{Subjective evaluation results for different video attribution methods by the pointing game metric.} Results on the metric are measured by percentage. The higher value means better performance in attribution.}
	\label{tab:quan_spg}
    \begin{tabularx}{\linewidth}{@{}l *{4}{Y}@{}}
    \toprule
    Methods
    & \multicolumn{2}{c}{UCF101-24}  
    & \multicolumn{2}{c}{EPIC-Object}\\
    \cmidrule(lr){2-3} \cmidrule(l){4-5}
    & R(2+1)D & R50L & R(2+1)D & R50L \\
    \midrule
    G*I~\cite{linear_appr} & 15.7 & 13.9 & 3.3 & 4.7 \\
    G~\cite{zeiler2014visual} & 31.8 & 26.8 & 6.4 & 5.9 \\
    IG~\cite{integrate_grad_icml} & 40.9 & 33.2 & 7.1 & 6.1 \\
    SG~\cite{smoothgrad} & 42.3 & 39.6 & 5.7 & 7.1 \\
    SG2~\cite{roar} & 44.4 & 38.8 & 5.7 & 7.3 \\
    BIG~\cite{blur_ig} & 39.3 & 30.4 & 6.7 & 6.3 \\
    XRAI~\cite{xrai} & 46.7 & 46.9 & 8.1 & 8.1 \\
    \midrule
    EB~\cite{excitbp,excit_bp_rnn} & 43.0 & 39.3 & 6.5 & 7.6  \\
    GC~\cite{grad_cam} & 48.3 & 27.3 & 7.8 & 5.9 \\
    \midrule
    STEP & \textbf{56.1} & \textbf{53.1} & \textbf{9.4} & \textbf{8.4} \\
    \bottomrule
    \end{tabularx}
    \vspace{-1em}
\end{table}

We also quantitatively compare the attribution results generated by different methods, using the pointing game metric, which adopts manual annotations of bounding boxes that ground the regions related to the ground-truth label according to the human judgement. Specifically, the metric measures the percentage of importance maps whose maximum points fall into the annotation bounding boxes. Following \cite{excit_bp_rnn}, we set a tolerance radius of 7 pixels when calculating for the pointing game metric, \ie, one hit is recorded if a 7-pixel radial circle around the maximum point in an importance map intersects the ground-truth bounding box. 

The evaluation results on the two networks and two datasets are shown in \autoref{tab:quan_spg}. It can be seen that STEP gets the best performance in all cases. The measurements on EPIC-Object are obviously lower than that on UCF101-24. This is mainly because the ground-truth objects are small and global motions are fast in the EPIC-Kitchen dataset. As a result, the bounding box annotations for objects are not very accurate and it also becomes difficult for attribution maps to accurately localize these objects in video.

\subsection{Quantitative Comparison using Retrain-based Metric}

\begin{figure}[!b]
    \centering
    \vspace{-2em}
    \includegraphics[width=0.8\linewidth]{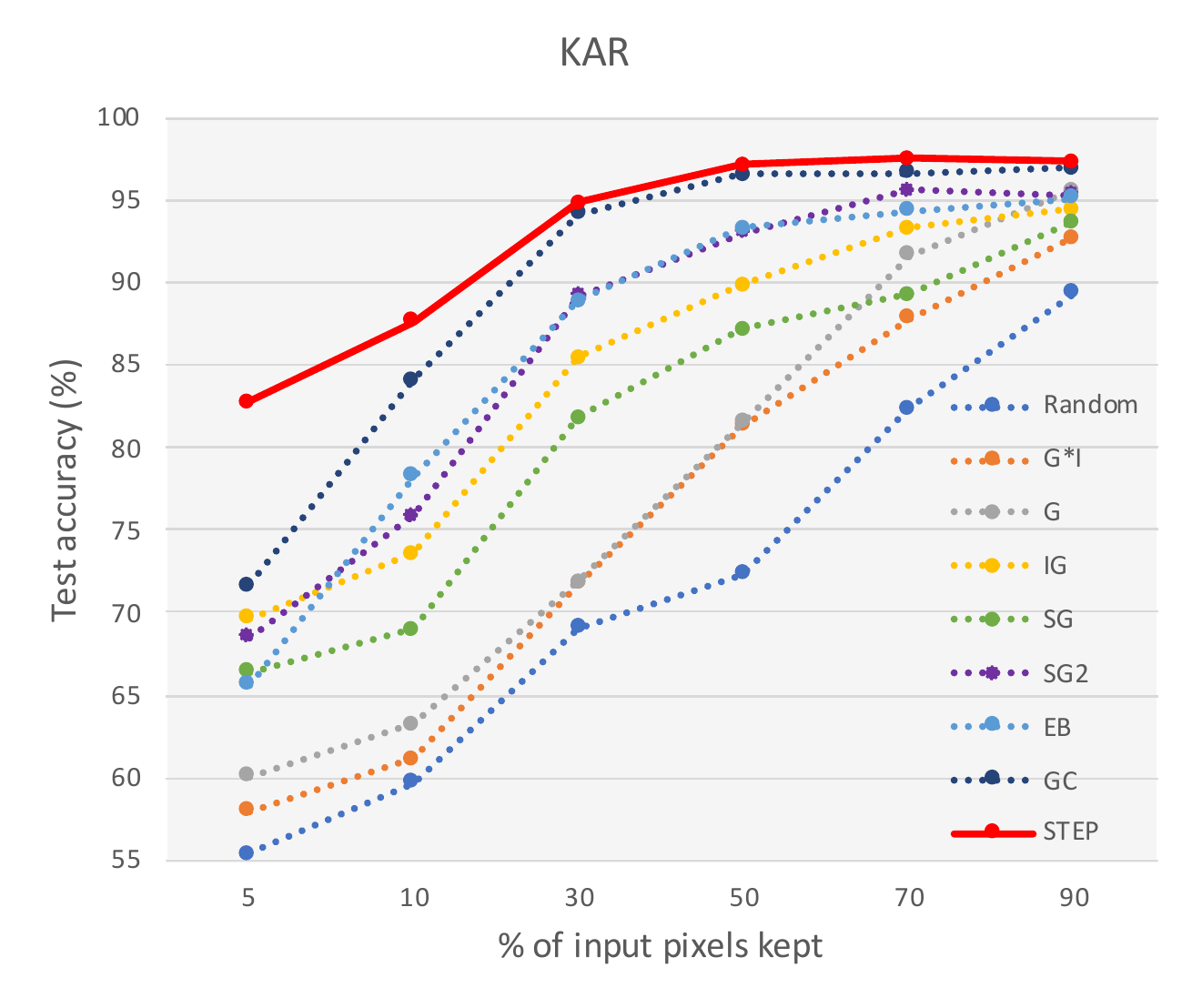}
    \vspace{-1em}
    \caption[]{\textbf{Objective evaluation results by the retrain-based metric KAR}. We use the perturbed UCF101-24 samples to retrain and test the R(2+1)D models. In these samples, only a part of the pixels that are considered contributive by the attribution maps is kept.}
    \label{fig:kar_comp_baseline}
\end{figure}

We also adopt the retrain-based metric~\cite{roar} to evaluate our method. Similar to the AUC-based metric, there are two different versions of retrain-based metrics (ROAR/KAR) due to the difference in perturbation operations (deletion/insertion). To be consistent with the AUC-based metric, we choose the KAR version based on the insertion operation to compare the performance of STEP against baseline methods. Since retrain-based metric needs to set multiple perturbation ratio values, and use the perturbed samples under each perturbation ratio value to train and test the model, it requires a lot of computational resources. We generated multiple training and test datasets on UCF101-24 by perturbing samples at the ratios of \{0.05, 0.1, 0.3, 0.5, 0.7, 0.9\} respectively, and apply them on the R(2+1)D models. For KAR, at each perturbation ratio, the attribution method that corresponds to a higher test accuracy is considered better, since this implies the attribution maps generated by the method can accurately locate the input regions that are discriminative to the model. The evaluation results are shown in \autoref{fig:kar_comp_baseline}. 

Under the small keeping ratio (5\% or 10\%), the test accuracy of our STEP method is significantly higher than that of baseline methods. Under large keeping ratios (no less than 30\%), the test accuracy values of STEP are still better than that of other baseline methods although the gaps with Grad-CAM are decreased. This indicates that the attribute maps generated by our STEP method can well locate the regions that occupy only a small proportion of the input but account truly significant for model discrimination.

\section{Conclusion}
In this paper, we shed light on the task of visually explaining video understanding networks by proposing a perturbation-based video attribution method: Spatio-Temporal Extremal Perturbation (STEP). The method adapts the extremal perturbation method for the video input and enhanced it with a new regularization term to smooth the perturbation results in both spatial and temporal dimensions. Instead of only utilizing the subjective metrics that rely on manual inspections or annotations, we incorporated the objective metrics to evaluate and compare different methods for video attribution. Our experiments indicated that different versions of an objective metric cannot come to a consensus in ranking a set of video attribution methods, which indicates the potential unreliability of some versions. Hence, we designed a new measurement for the AUC-based metrics to reveal and quantify their reliability. We experiment on two typical backbone networks (3D-CNNs \& CNN-RNN) for the video classification task and three datasets of Something-Something-V2 \& EPIC-Kitchens \& UCF101-24. For a comprehensive comparison, we also incorporated multiple significant attribution methods into the baseline, which were originally proposed as image attribution methods but adaptive for the video task. Both the objective and subjective evaluation results demonstrate that our proposed method can achieve competitive performance as well as maintain decent resolutions and temporal sensitivity on attribution results.

\ifCLASSOPTIONcaptionsoff
  \newpage
\fi

\bibliographystyle{IEEEtran}
\bibliography{mainbib}

\begin{thebibliography}{10}
\providecommand{\url}[1]{#1}
\csname url@samestyle\endcsname
\providecommand{\newblock}{\relax}
\providecommand{\bibinfo}[2]{#2}
\providecommand{\BIBentrySTDinterwordspacing}{\spaceskip=0pt\relax}
\providecommand{\BIBentryALTinterwordstretchfactor}{4}
\providecommand{\BIBentryALTinterwordspacing}{\spaceskip=\fontdimen2\font plus
\BIBentryALTinterwordstretchfactor\fontdimen3\font minus
  \fontdimen4\font\relax}
\providecommand{\BIBforeignlanguage}[2]{{%
\expandafter\ifx\csname l@#1\endcsname\relax
\typeout{** WARNING: IEEEtran.bst: No hyphenation pattern has been}%
\typeout{** loaded for the language `#1'. Using the pattern for}%
\typeout{** the default language instead.}%
\else
\language=\csname l@#1\endcsname
\fi
#2}}
\providecommand{\BIBdecl}{\relax}
\BIBdecl

\bibitem{two_stream}
K.~Simonyan and A.~Zisserman, ``Two-stream convolutional networks for action
  recognition in videos,'' in \emph{Advances in Neural Information Processing
  Systems (NeurIPS)}, 2014, pp. 568--576.

\bibitem{i3d}
J.~Carreira and A.~Zisserman, ``Quo vadis, action recognition? a new model and
  the kinetics dataset,'' in \emph{IEEE Conference on Computer Vision and
  Pattern Recognition (CVPR)}, 2017, pp. 6299--6308.

\bibitem{tsn}
L.~Wang, Y.~Xiong, Z.~Wang, Y.~Qiao, D.~Lin, X.~Tang, and L.~Van~Gool,
  ``Temporal segment networks: Towards good practices for deep action
  recognition,'' in \emph{European Conference on Computer Vision (ECCV)}.\hskip
  1em plus 0.5em minus 0.4em\relax Springer, 2016, pp. 20--36.

\bibitem{tcsvt_action_recog1}
N.~A. Tu, T.~Huynh-The, K.~U. Khan, and Y.-K. Lee, ``Ml-hdp: a hierarchical
  bayesian nonparametric model for recognizing human actions in video,''
  \emph{IEEE Transactions on Circuits and Systems for Video Technology
  (TCSVT)}, vol.~29, no.~3, pp. 800--814, 2018.

\bibitem{dense_videocaption}
R.~Krishna, K.~Hata, F.~Ren, L.~Fei-Fei, and J.~Carlos~Niebles,
  ``Dense-captioning events in videos,'' in \emph{IEEE International Conference
  on Computer Vision (ICCV)}, 2017, pp. 706--715.

\bibitem{videocap_transf}
L.~Zhou, Y.~Zhou, J.~J. Corso, R.~Socher, and C.~Xiong, ``End-to-end dense
  video captioning with masked transformer,'' in \emph{IEEE Conference on
  Computer Vision and Pattern Recognition (CVPR)}, 2018, pp. 8739--8748.

\bibitem{tcsvt_video_caption}
N.~Xu, A.-A. Liu, Y.~Wong, Y.~Zhang, W.~Nie, Y.~Su, and M.~Kankanhalli,
  ``Dual-stream recurrent neural network for video captioning,'' \emph{IEEE
  Transactions on Circuits and Systems for Video Technology (TCSVT)}, vol.~29,
  no.~8, pp. 2482--2493, 2018.

\bibitem{vqa}
S.~Antol, A.~Agrawal, J.~Lu, M.~Mitchell, D.~Batra, C.~Lawrence~Zitnick, and
  D.~Parikh, ``Vqa: Visual question answering,'' in \emph{IEEE International
  Conference on Computer Vision (ICCV)}, 2015, pp. 2425--2433.

\bibitem{memory_vqa}
J.~Gao, R.~Ge, K.~Chen, and R.~Nevatia, ``Motion-appearance co-memory networks
  for video question answering,'' in \emph{IEEE Conference on Computer Vision
  and Pattern Recognition (CVPR)}, 2018, pp. 6576--6585.

\bibitem{tcsvt_vqa}
Y.-C. Wu and J.-C. Yang, ``A robust passage retrieval algorithm for video
  question answering,'' \emph{IEEE Transactions on Circuits and Systems for
  Video Technology (TCSVT)}, vol.~18, no.~10, pp. 1411--1421, 2008.

\bibitem{tcsvt_saliency_prediction1}
K.~Zhang and Z.~Chen, ``Video saliency prediction based on spatial-temporal
  two-stream network,'' \emph{IEEE Transactions on Circuits and Systems for
  Video Technology (TCSVT)}, vol.~29, no.~12, pp. 3544--3557, 2019.

\bibitem{tcsvt_saliency_prediction2}
Z.~Wu, L.~Su, and Q.~Huang, ``Learning coupled convolutional networks fusion
  for video saliency prediction,'' \emph{IEEE Transactions on Circuits and
  Systems for Video Technology (TCSVT)}, vol.~29, no.~10, pp. 2960--2971, 2019.

\bibitem{tcsvt_saliency_detection1}
Z.~Liu, X.~Zhang, S.~Luo, and O.~Le~Meur, ``Superpixel-based spatiotemporal
  saliency detection,'' \emph{IEEE Transactions on Circuits and Systems for
  Video Technology (TCSVT)}, vol.~24, no.~9, pp. 1522--1540, 2014.

\bibitem{tcsvt_saliency_detection2}
R.~Cong, J.~Lei, H.~Fu, M.-M. Cheng, W.~Lin, and Q.~Huang, ``Review of visual
  saliency detection with comprehensive information,'' \emph{IEEE Transactions
  on Circuits and Systems for Video Technology}, vol.~29, no.~10, pp.
  2941--2959, 2019.

\bibitem{zeiler2014visual}
M.~D. Zeiler and R.~Fergus, ``Visualizing and understanding convolutional
  networks,'' in \emph{European Conference on Computer Vision (ECCV)}, 2014,
  pp. 818--833.

\bibitem{meaningful_perturb}
R.~C. Fong and A.~Vedaldi, ``Interpretable explanations of black boxes by
  meaningful perturbation,'' in \emph{IEEE International Conference on Computer
  Vision (ICCV)}, 2017, pp. 3429--3437.

\bibitem{baehrens2010explain}
D.~Baehrens, T.~Schroeter, S.~Harmeling, M.~Kawanabe, K.~Hansen, and K.-R.
  M{\~A}{\v{z}}ller, ``How to explain individual classification decisions,''
  \emph{Journal of Machine Learning Research (JMLR)}, vol.~11, no. Jun, pp.
  1803--1831, 2010.

\bibitem{grad}
K.~Simonyan, A.~Vedaldi, and A.~Zisserman, ``Deep inside convolutional
  networks: Visualising image classification models and saliency maps,''
  \emph{arXiv preprint arXiv:1312.6034}, 2013.

\bibitem{this_looks_like}
C.~Chen, O.~Li, D.~Tao, A.~Barnett, C.~Rudin, and J.~K. Su, ``This looks like
  that: Deep learning for interpretable image recognition,'' in \emph{Advances
  in Neural Information Processing Systems}, vol.~32, 2019.

\bibitem{interp_cnns}
Q.~{Zhang}, Y.~N. {Wu}, and S.~{Zhu}, ``Interpretable convolutional neural
  networks,'' in \emph{IEEE/CVF Conference on Computer Vision and Pattern
  Recognition (CVPR)}, 2018, pp. 8827--8836.

\bibitem{guided_zoom}
S.~A. {Bargal}, A.~{Zunino}, V.~{Petsiuk}, J.~{Zhang}, K.~{Saenko},
  V.~{Murino}, and S.~{Sclaroff}, ``Guided zoom: Zooming into network evidence
  to refine fine-grained model decisions,'' \emph{IEEE Transactions on Pattern
  Analysis and Machine Intelligence (PAMI)}, 2021.

\bibitem{integrate_grad_icml}
M.~Sundararajan, A.~Taly, and Q.~Yan, ``Axiomatic attribution for deep
  networks,'' in \emph{International Conference on Machine Learning (ICML)},
  2017, pp. 3319--3328.

\bibitem{smoothgrad}
D.~Smilkov, N.~Thorat, B.~Kim, F.~Vi{\'e}gas, and M.~Wattenberg, ``Smoothgrad:
  removing noise by adding noise,'' \emph{arXiv preprint arXiv:1706.03825},
  2017.

\bibitem{lrp}
S.~Bach, A.~Binder, G.~Montavon, F.~Klauschen, K.-R. M{\"u}ller, and W.~Samek,
  ``On pixel-wise explanations for non-linear classifier decisions by
  layer-wise relevance propagation,'' \emph{PloS one}, vol.~10, no.~7, 2015.

\bibitem{excitbp}
J.~Zhang, S.~A. Bargal, Z.~Lin, J.~Brandt, X.~Shen, and S.~Sclaroff, ``Top-down
  neural attention by excitation backprop,'' \emph{International Journal of
  Computer Vision (IJCV)}, vol. 126, no.~10, pp. 1084--1102, 2018.

\bibitem{grad_cam}
R.~R. Selvaraju, M.~Cogswell, A.~Das, R.~Vedantam, D.~Parikh, and D.~Batra,
  ``Grad-cam: Visual explanations from deep networks via gradient-based
  localization,'' in \emph{IEEE International Conference on Computer Vision
  (ICCV)}, 2017, pp. 618--626.

\bibitem{excit_bp_rnn}
S.~Adel~Bargal, A.~Zunino, D.~Kim, J.~Zhang, V.~Murino, and S.~Sclaroff,
  ``Excitation backprop for rnns,'' in \emph{IEEE Conference on Computer Vision
  and Pattern Recognition (CVPR)}, 2018, pp. 1440--1449.

\bibitem{saliency_tubes}
A.~Stergiou, G.~Kapidis, G.~Kalliatakis, C.~Chrysoulas, R.~Veltkamp, and
  R.~Poppe, ``Saliency tubes: Visual explanations for spatio-temporal
  convolutions,'' in \emph{IEEE International Conference on Image Processing
  (ICIP)}, 2019, pp. 1830--1834.

\bibitem{wacv_version}
Z.~Li, W.~Wang, Z.~Li, Y.~Huang, and Y.~Sato, ``Towards visually explaining
  video understanding networks with perturbation,'' in \emph{IEEE/CVF Winter
  Conference on Applications of Computer Vision (WACV)}, 2021, pp. 1120--1129.

\bibitem{linear_appr}
P.-J. Kindermans, K.~Sch{\"u}tt, K.-R. M{\"u}ller, and S.~D{\"a}hne,
  ``Investigating the influence of noise and distractors on the interpretation
  of neural networks,'' \emph{arXiv preprint arXiv:1611.07270}, 2016.

\bibitem{guided_bp_iclr}
J.~Springenberg, A.~Dosovitskiy, T.~Brox, and M.~Riedmiller, ``Striving for
  simplicity: The all convolutional net,'' in \emph{ICLR (workshop track)},
  2015.

\bibitem{deeplift}
A.~Shrikumar, P.~Greenside, and A.~Kundaje, ``Learning important features
  through propagating activation differences,'' in \emph{International
  Conference on Machine Learning (ICML)}, 2017, pp. 3145--3153.

\bibitem{roar}
S.~Hooker, D.~Erhan, P.-J. Kindermans, and B.~Kim, ``A benchmark for
  interpretability methods in deep neural networks,'' in \emph{Advances in
  Neural Information Processing Systems (NeurIPS)}, 2019, pp. 9737--9748.

\bibitem{xrai}
A.~Kapishnikov, T.~Bolukbasi, F.~Vi{\'e}gas, and M.~Terry, ``Xrai: Better
  attributions through regions,'' in \emph{IEEE International Conference on
  Computer Vision (ICCV)}, 2019, pp. 4948--4957.

\bibitem{blur_ig}
S.~Xu, S.~Venugopalan, and M.~Sundararajan, ``Attribution in scale and space,''
  in \emph{IEEE/CVF Conference on Computer Vision and Pattern Recognition
  (CVPR)}, 2020, pp. 9680--9689.

\bibitem{cam}
B.~Zhou, A.~Khosla, A.~Lapedriza, A.~Oliva, and A.~Torralba, ``Learning deep
  features for discriminative localization,'' in \emph{IEEE Conference on
  Computer Vision and Pattern Recognition (CVPR)}, 2016, pp. 2921--2929.

\bibitem{grad_cam_plusplus}
A.~Chattopadhay, A.~Sarkar, P.~Howlader, and V.~N. Balasubramanian,
  ``Grad-cam++: Generalized gradient-based visual explanations for deep
  convolutional networks,'' in \emph{IEEE Winter Conference on Applications of
  Computer Vision (WACV)}, 2018, pp. 839--847.

\bibitem{score_cam}
H.~Wang, Z.~Wang, M.~Du, F.~Yang, Z.~Zhang, S.~Ding, P.~Mardziel, and X.~Hu,
  ``Score-cam: Score-weighted visual explanations for convolutional neural
  networks,'' in \emph{IEEE/CVF Conference on Computer Vision and Pattern
  Recognition Workshops (CVPRW)}, 2020, pp. 24--25.

\bibitem{lime}
M.~T. Ribeiro, S.~Singh, and C.~Guestrin, ``" why should i trust you?"
  explaining the predictions of any classifier,'' in \emph{International
  Conference on Knowledge Discovery and Data Mining (KDD)}, 2016, pp.
  1135--1144.

\bibitem{rise}
V.~Petsiuk, A.~Das, and K.~Saenko, ``Rise: Randomized input sampling for
  explanation of black-box models,'' \emph{In British Machine Vision Conference
  (BMVC)}, 2018.

\bibitem{igos}
Z.~Qi, S.~Khorram, and F.~Li, ``Visualizing deep networks by optimizing with
  integrated gradients,'' in \emph{IEEE Conference on Computer Vision and
  Pattern Recognition Workshops (CVPRW)}, 2019, pp. 1--4.

\bibitem{fgvis}
J.~Wagner, J.~M. Kohler, T.~Gindele, L.~Hetzel, J.~T. Wiedemer, and S.~Behnke,
  ``Interpretable and fine-grained visual explanations for convolutional neural
  networks,'' in \emph{IEEE Conference on Computer Vision and Pattern
  Recognition (CVPR)}, 2019, pp. 9097--9107.

\bibitem{extremal_perturb}
R.~Fong, M.~Patrick, and A.~Vedaldi, ``Understanding deep networks via extremal
  perturbations and smooth masks,'' in \emph{IEEE International Conference on
  Computer Vision (ICCV)}, 2019, pp. 2950--2958.

\bibitem{tcsvt_action_recognition}
T.~V. Nguyen, Z.~Song, and S.~Yan, ``Stap: Spatial-temporal attention-aware
  pooling for action recognition,'' \emph{IEEE Transactions on Circuits and
  Systems for Video Technology (TCSVT)}, vol.~25, no.~1, pp. 77--86, 2015.

\bibitem{norm_grad}
S.-A. Rebuffi, R.~Fong, X.~Ji, and A.~Vedaldi, ``There and back again:
  Revisiting backpropagation saliency methods,'' in \emph{IEEE Conference on
  Computer Vision and Pattern Recognition (CVPR)}, 2020, pp. 8839--8848.

\bibitem{realtime_saliency}
P.~Dabkowski and Y.~Gal, ``Real time image saliency for black box
  classifiers,'' in \emph{Advances in Neural Information Processing Systems
  (NeurIPS)}, 2017, pp. 6967--6976.

\bibitem{devnet}
C.~Gan, N.~Wang, Y.~Yang, D.-Y. Yeung, and A.~G. Hauptmann, ``Devnet: A deep
  event network for multimedia event detection and evidence recounting,'' in
  \emph{IEEE Conference on Computer Vision and Pattern Recognition (CVPR)},
  2015, pp. 2568--2577.

\bibitem{lrp_video}
C.~J. Anders, G.~Montavon, W.~Samek, and K.-R. M{\"u}ller, ``Understanding
  patch-based learning of video data by explaining predictions,'' in
  \emph{Explainable AI: Interpreting, Explaining and Visualizing Deep
  Learning}.\hskip 1em plus 0.5em minus 0.4em\relax Springer, 2019, pp.
  297--309.

\bibitem{sth_sth}
R.~Goyal, S.~E. Kahou, V.~Michalski, J.~Materzynska, S.~Westphal, H.~Kim,
  V.~Haenel, I.~Fruend, P.~Yianilos, M.~Mueller-Freitag \emph{et~al.}, ``The"
  something something" video database for learning and evaluating visual common
  sense,'' in \emph{IEEE International Conference on Computer Vision (ICCV)},
  vol.~1, no.~4, 2017, p.~5.

\bibitem{stergiou2019class}
A.~Stergiou, G.~Kapidis, G.~Kalliatakis, C.~Chrysoulas, R.~Poppe, and
  R.~Veltkamp, ``Class feature pyramids for video explanation,'' \emph{arXiv
  preprint arXiv:1909.08611}, 2019.

\bibitem{aopc}
W.~Samek, A.~Binder, G.~Montavon, S.~Lapuschkin, and K.-R. M{\"u}ller,
  ``Evaluating the visualization of what a deep neural network has learned,''
  \emph{IEEE Transactions on Neural Networks and Learning Systems}, vol.~28,
  no.~11, pp. 2660--2673, 2016.

\bibitem{irof}
L.~Rieger and L.~K. Hansen, ``Irof: a low resource evaluation metric for
  explanation methods,'' \emph{arXiv preprint arXiv:2003.08747}, 2020.

\bibitem{slic}
R.~Achanta, A.~Shaji, K.~Smith, A.~Lucchi, P.~Fua, and S.~S{\"u}sstrunk, ``Slic
  superpixels compared to state-of-the-art superpixel methods,'' \emph{IEEE
  Transactions on Pattern Analysis and Machine Intelligence (PAMI)}, vol.~34,
  no.~11, pp. 2274--2282, 2012.

\bibitem{adversarial_unrecognizable}
A.~Nguyen, J.~Yosinski, and J.~Clune, ``Deep neural networks are easily fooled:
  High confidence predictions for unrecognizable images,'' in \emph{IEEE
  Conference on Computer Vision and Pattern Recognition (CVPR)}, 2015, pp.
  427--436.

\bibitem{adversarial_noised}
I.~J. Goodfellow, J.~Shlens, and C.~Szegedy, ``Explaining and harnessing
  adversarial examples,'' \emph{arXiv preprint arXiv:1412.6572}, 2014.

\bibitem{adversarial}
A.~Kurakin, I.~Goodfellow, and S.~Bengio, ``Adversarial examples in the
  physical world,'' \emph{arXiv preprint arXiv:1607.02533}, 2016.

\bibitem{adversarial_intriguing}
C.~Szegedy, W.~Zaremba, I.~Sutskever, J.~Bruna, D.~Erhan, I.~Goodfellow, and
  R.~Fergus, ``Intriguing properties of neural networks,'' \emph{arXiv preprint
  arXiv:1312.6199}, 2013.

\bibitem{generative_adversarial}
I.~Goodfellow, J.~Pouget-Abadie, M.~Mirza, B.~Xu, D.~Warde-Farley, S.~Ozair,
  A.~Courville, and Y.~Bengio, ``Generative adversarial nets,'' in
  \emph{Advances in Neural Information Processing Systems (NeurIPS)}, 2014, pp.
  2672--2680.

\bibitem{resnet}
K.~He, X.~Zhang, S.~Ren, and J.~Sun, ``Deep residual learning for image
  recognition,'' in \emph{IEEE Conference on Computer Vision and Pattern
  Recognition (CVPR)}, 2016, pp. 770--778.

\bibitem{lstm}
S.~Hochreiter and J.~Schmidhuber, ``Long short-term memory,'' \emph{Neural
  Computation}, vol.~9, no.~8, pp. 1735--1780, 1997.

\bibitem{r2plus1d}
D.~Tran, H.~Wang, L.~Torresani, J.~Ray, Y.~LeCun, and M.~Paluri, ``A closer
  look at spatiotemporal convolutions for action recognition,'' in \emph{IEEE
  conference on Computer Vision and Pattern Recognition (CVPR)}, 2018, pp.
  6450--6459.

\bibitem{ucf101}
K.~Soomro, A.~R. Zamir, and M.~Shah, ``Ucf101: A dataset of 101 human actions
  classes from videos in the wild,'' \emph{arXiv preprint arXiv:1212.0402},
  2012.

\bibitem{epic}
D.~Damen, H.~Doughty, G.~Maria~Farinella, S.~Fidler, A.~Furnari, E.~Kazakos,
  D.~Moltisanti, J.~Munro, T.~Perrett, W.~Price \emph{et~al.}, ``Scaling
  egocentric vision: The epic-kitchens dataset,'' in \emph{European Conference
  on Computer Vision (ECCV)}, 2018, pp. 720--736.

\bibitem{dance_in_mall}
J.~Choi, C.~Gao, J.~C. Messou, and J.-B. Huang, ``Why can't i dance in the
  mall? learning to mitigate scene bias in action recognition,'' in
  \emph{Advances in Neural Information Processing Systems (NeurIPS)}, 2019, pp.
  853--865.

\bibitem{diving48}
Y.~Li, Y.~Li, and N.~Vasconcelos, ``Resound: Towards action recognition without
  representation bias,'' in \emph{European Conference on Computer Vision
  (ECCV)}, 2018, pp. 513--528.

\end{thebibliography}

\newcommand{\photowidth}{1.00in}
\newcommand{\photoheight}{1.25in}
\newcommand{\biosep}{\vspace{-4em}}

\biosep
\begin{IEEEbiography}
[{\includegraphics[width=\photowidth,height=\photoheight,clip,keepaspectratio]{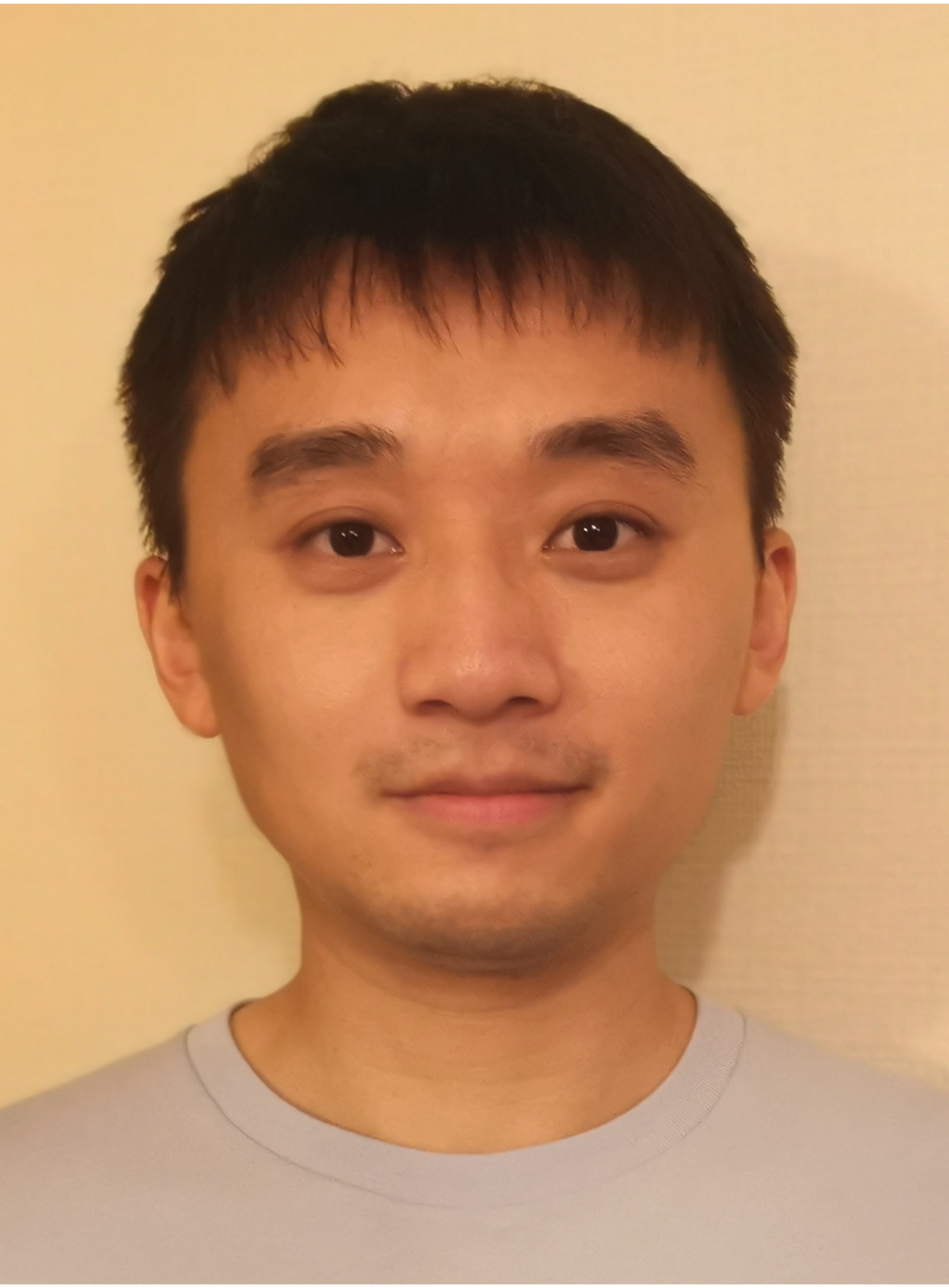}}]{Zhenqiang Li} received his B.S. degree from Huazhong University of Science and Technology, Wuhan, China, in 2016 and received M.S. degree in information science and technology from The University of Tokyo in 2019. He is now continuing for Ph. D. in the University of Tokyo. His research interests include video understanding, first-person vision and skill assessment.
\end{IEEEbiography}
\biosep
\begin{IEEEbiography}
[{\includegraphics[width=\photowidth,height=\photoheight,clip,keepaspectratio]{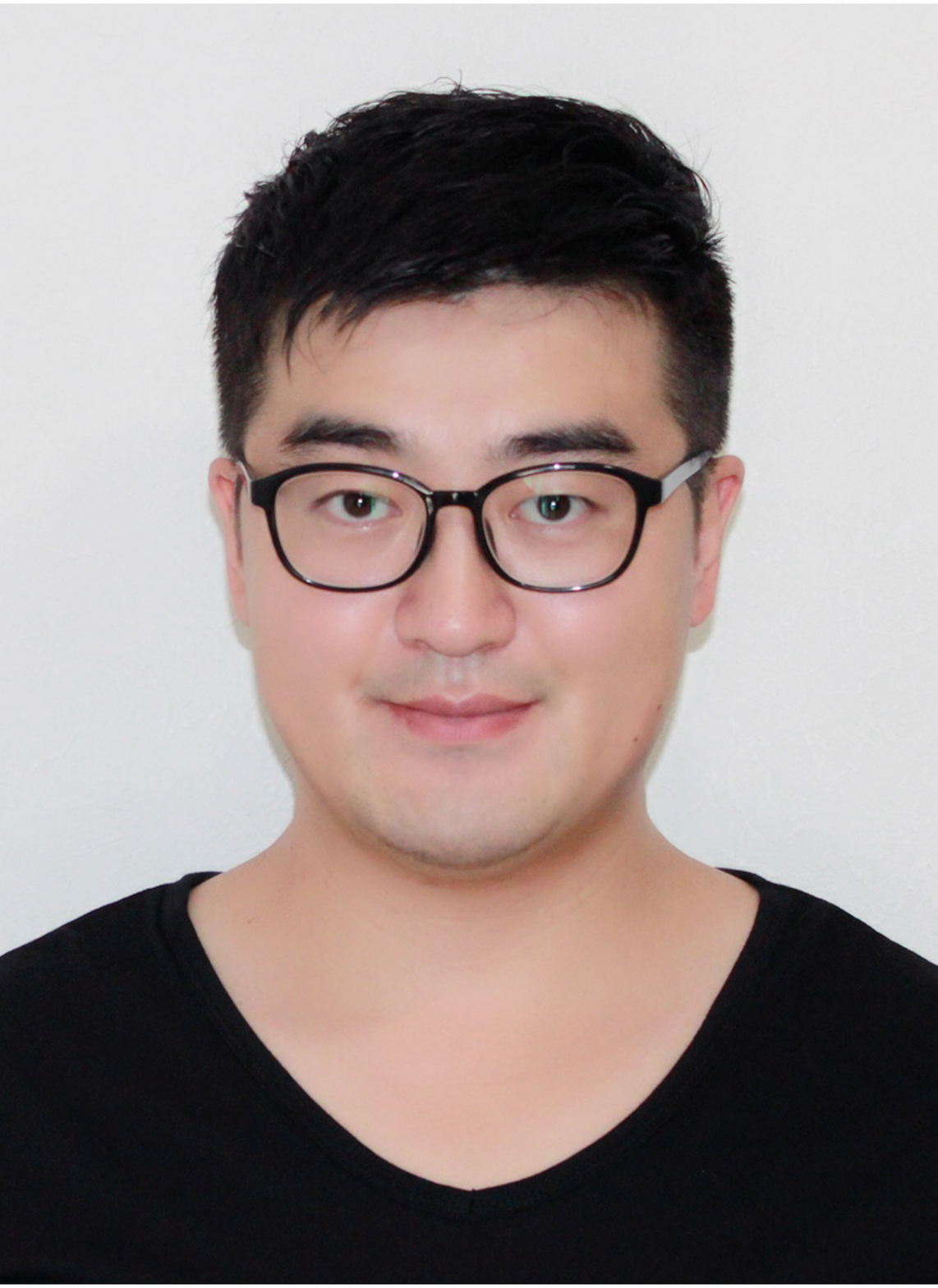}}]{Weimin Wang} received the B.S. degree from Shanghai Jiao Tong University in 2009, the M.S. degree from Osaka University in 2012, and the Ph.D. degree from Nagoya University in 2017. He was a researcher at AIST, Japan, from 2018 to 2021. He is currently an associate professor with the DUT-RU International School of Information Science and Engineering, Dalian University of Technology. His research interests include computer vision, multi-modal data fusion, scene understanding.
\end{IEEEbiography}
\biosep
\begin{IEEEbiography}
[{\includegraphics[width=\photowidth,height=\photoheight,clip,keepaspectratio]{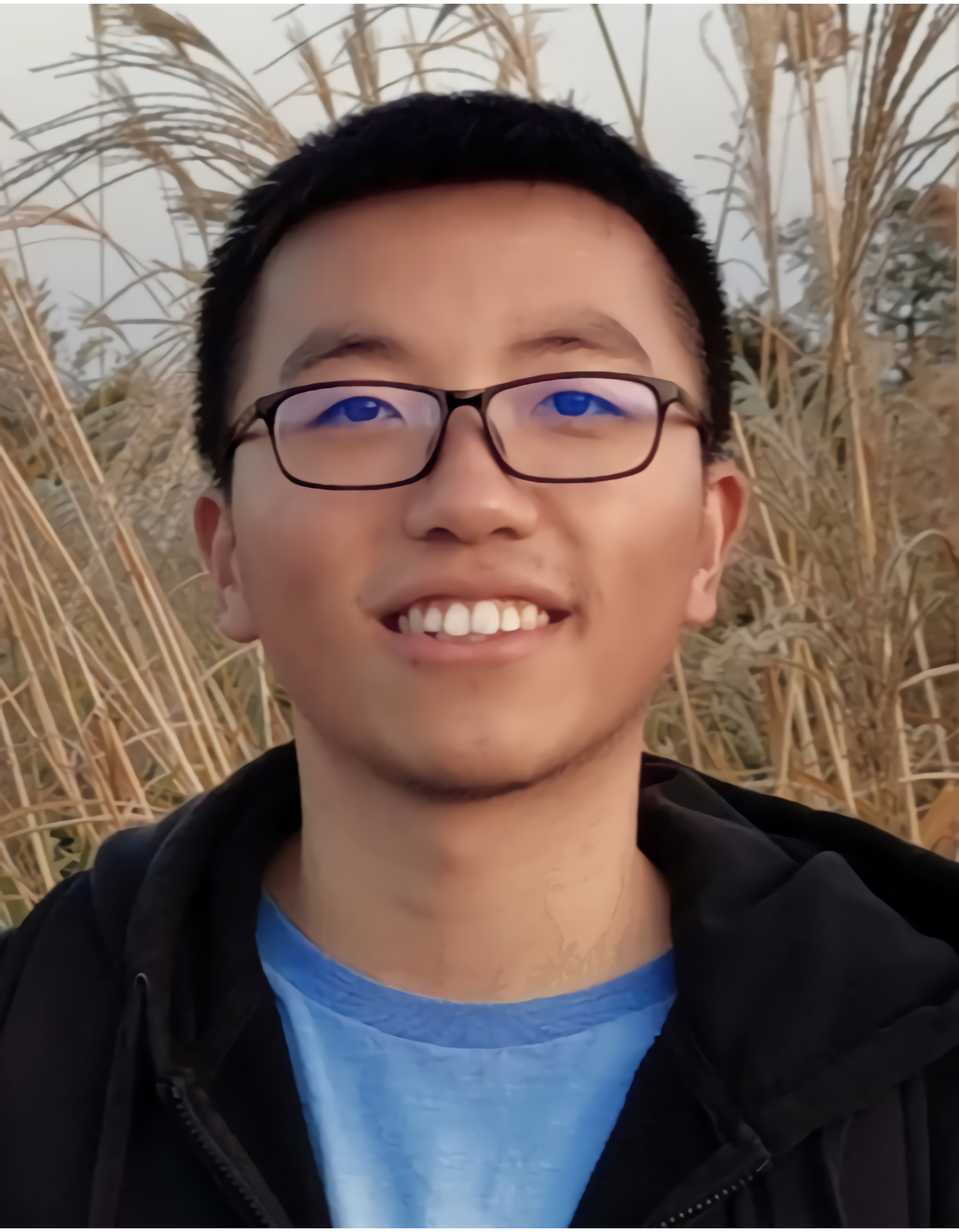}}]{Zuoyue Li} received his B.Eng. degree from Zhejiang University, Hangzhou, China, in 2015 and received his M.Sc. degree from ETH Zürich, Switzerland in 2018. He is now a PhD student at ETH Zürich and has research interests in 3D vision, scene understanding and remote sensing
\end{IEEEbiography}
\biosep
\begin{IEEEbiography}
[{\includegraphics[width=\photowidth,height=\photoheight,clip,keepaspectratio]{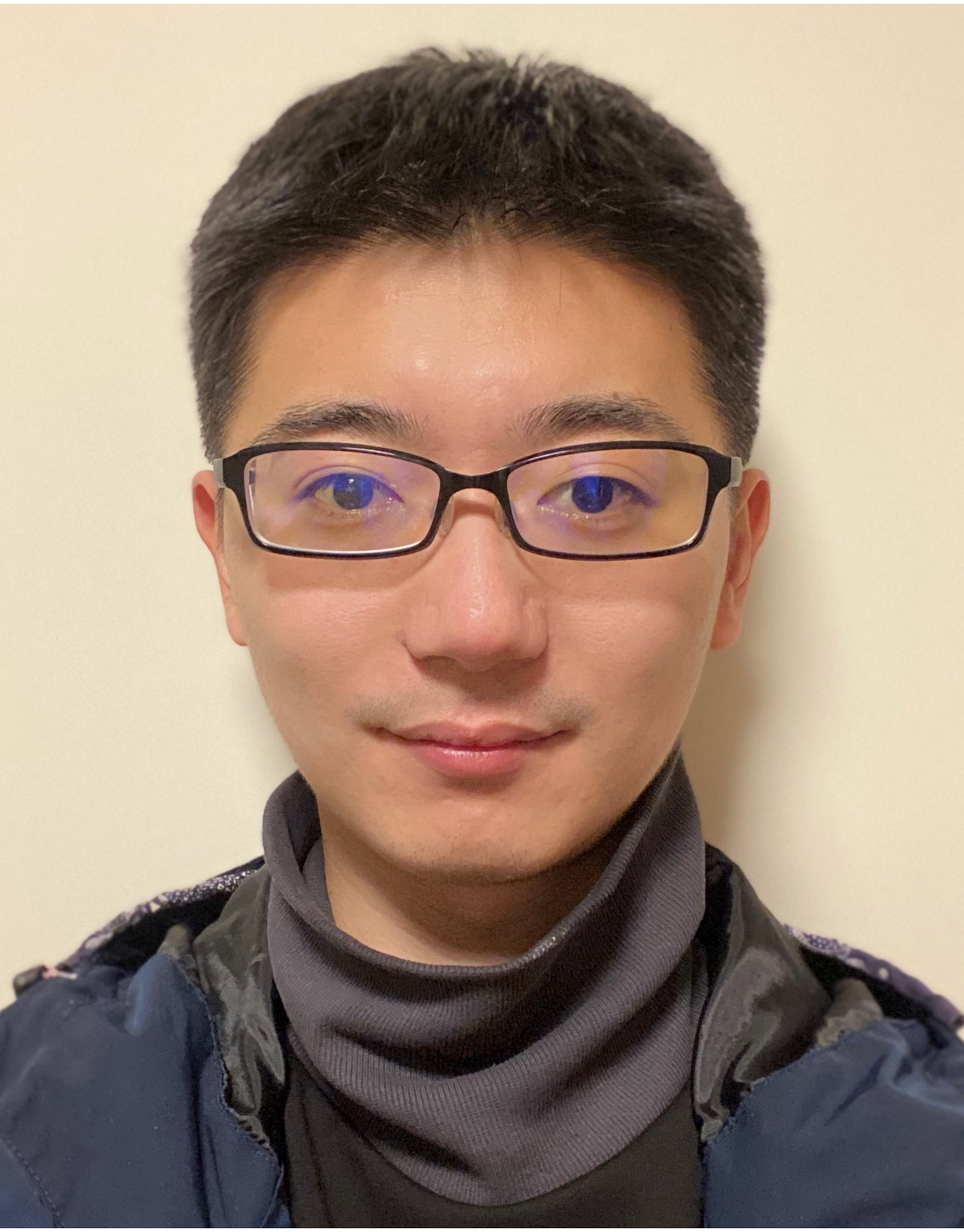}}]
{Yifei Huang} received his B.S. degree in automation from Shanghai Jiao Tong University, Shanghai, China in 2015 and received M.S., Ph.D. degrees in information science and technology from The University of Tokyo, Tokyo, Japan, in 2018 and 2021. He is now a reseach associate in the Institute of Industrial Science, the University of Tokyo. His research interests include first-person vision and their applications, medical image processing, and robotics.
\end{IEEEbiography}
\biosep
\begin{IEEEbiography}
[{\includegraphics[width=\photowidth,height=\photoheight,clip,keepaspectratio]{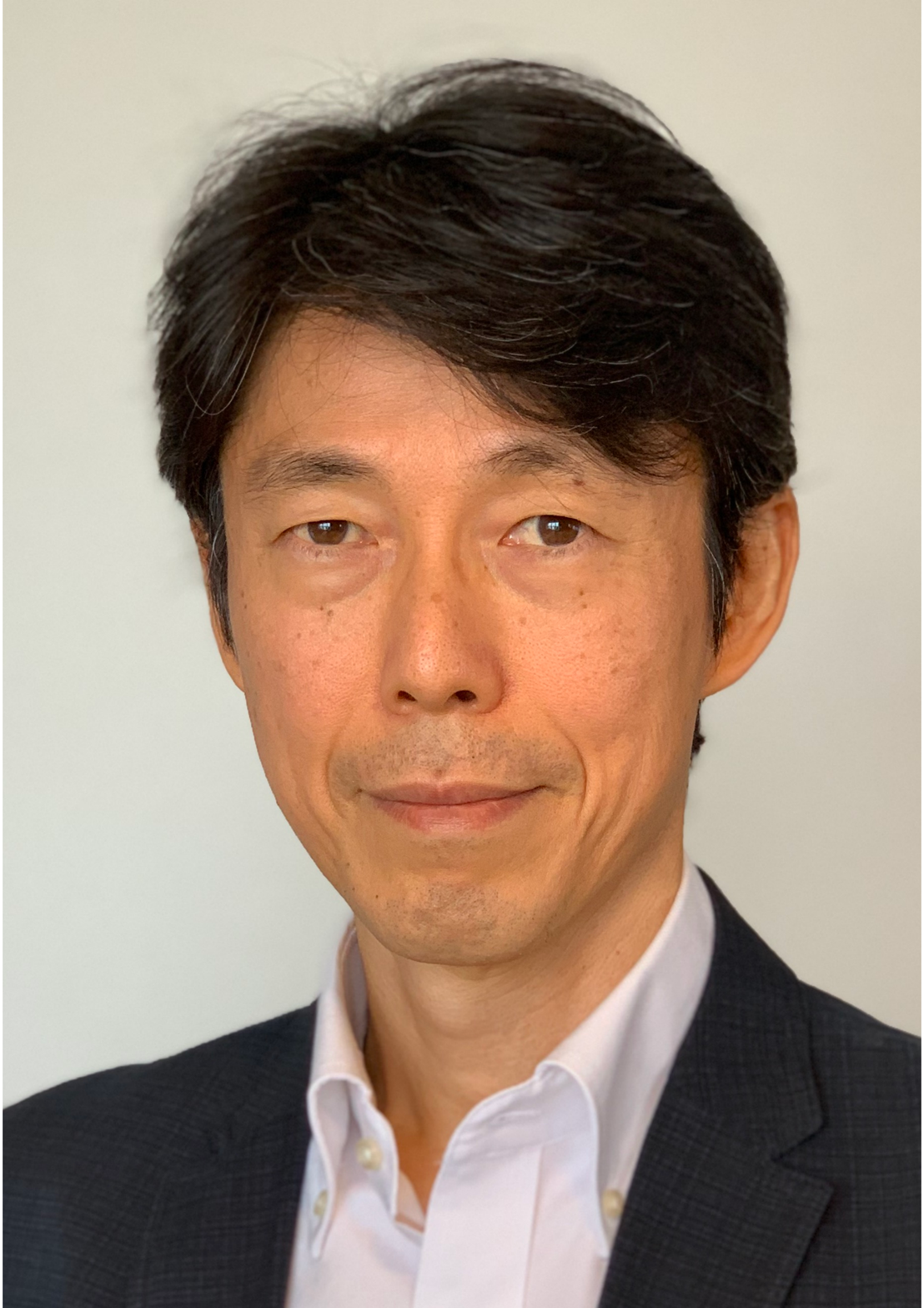}}]{Yoichi Sato} is a professor at Institute of Industrial Science, the University of Tokyo. He received his B.S. degree from the University of Tokyo in 1990, and his MS and Ph.D. degrees in robotics from School of Computer Science, Carnegie Mellon University in 1993 and 1997. His research interests include physics-based vision, reflectance analysis, first-person vision, and gaze sensing and analysis. He served/is serving in several conference organization and journal editorial roles including IEEE Transactions on Pattern Analysis and Machine Intelligence, International Journal of Computer Vision, Computer Vision and Image Understanding, CVPR 2023 General Co-Chair, ICCV 2021 Program Co-Chair, ACCV 2018 General Co-Chair, ACCV 2016 Program Co-Chair, and ECCV 2012 Program Co-Chair.
\end{IEEEbiography}

\end{document}